\documentclass[table]{article}

\usepackage[preprint]{neurips_2020}
\usepackage{natbib}

\usepackage[utf8]{inputenc} 
\usepackage[T1]{fontenc}    
\usepackage{hyperref}       
\usepackage{url}            
\usepackage{booktabs}       
\usepackage{amsfonts}       
\usepackage{nicefrac}       
\usepackage{microtype}      
\usepackage{sidecap}
\usepackage[table]{xcolor}

\usepackage{multirow}
\usepackage{graphicx}
\usepackage{subfigure}
\usepackage{dcolumn}
\usepackage{amsmath, amssymb, mathtools, bm, amsthm}
\usepackage{booktabs} 
\usepackage{todonotes}
\usepackage{hyperref}
\usepackage{algorithm}
\usepackage{algorithmic}
\usepackage{bm}
\usepackage{cleveref}
\usepackage{caption}
\usepackage{minitoc}
\usepackage[toc,page,header]{appendix}

\newcommand{\modelname}{{\textsc{scoff}}}
\newcommand{\OF}{{\textsc{of}}}
\newcommand{\OFs}{{\textsc{of}s}}

\newcommand{\va}{{\bm{a}}}

\newcommand{\vh}{{\bm{h}}}
\newcommand{\ve}{{\bm{e}}}

\newcommand{\vc}{{\bm{c}}}

\newcommand{\vq}{{\bm{q}}}
\newcommand{\vx}{{\bm{x}}}
\newcommand{\vv}{{\bm{v}}}
\newcommand{\vz}{{\bm{z}}}

\newcommand{\vW}{{\bm{W}}}

\newcommand{\vkappa}{{\bm{\kappa}}}

\newcommand{\vtheta}{{\mbox{\boldmath\ensuremath{\theta}}}}

\usepackage{pifont}

\newcommand{\highlight}[1]{\colorbox{blue!10}{#1}}

\newcommand{\bull}{{\tiny $\bullet~~$ }}

\title{Object Files and Schemata: Factorizing Declarative and Procedural Knowledge in Dynamical Systems}

\author{Anirudh Goyal\textsuperscript{1}, 
Alex Lamb \textsuperscript{1},
Phanideep Gampa \textsuperscript{1, 2},
Philippe Beaudoin \textsuperscript{3},
Sergey Levine \textsuperscript{4}, \\
\textbf{Charles Blundell \textsuperscript{5},
Yoshua Bengio \textsuperscript{1},
Michael Mozer \textsuperscript{6}}}

\DeclareUnicodeCharacter{2212}{-}

\begin{document}

\let\footnote\relax\footnotetext{\textsuperscript{1} Mila, University of Montreal, \textsuperscript{2} IIT BHU, Varanasi, \textsuperscript{3} ElementAI, \textsuperscript{4} UC Berkeley, \textsuperscript{5} Deepmind, \textsuperscript{6} Google Research, Brain Team,
Corresponding author: \texttt{anirudhgoyal9119@gmail.com}
}

\maketitle

\begin{abstract}
Modeling a structured, dynamic environment like a video game requires keeping track of the objects and their states (\emph{declarative} knowledge) as well as predicting how objects behave (\emph{procedural} knowledge). Black-box models with a monolithic hidden state often fail to apply procedural knowledge consistently and uniformly, i.e., they lack \emph{systematicity}. For example, in a video game, correct prediction of one enemy's trajectory does not ensure correct prediction of another's. We address this issue via an architecture that factorizes declarative and procedural knowledge and that imposes modularity within each form of knowledge. The architecture consists of active modules called \emph{object files} that maintain the state of a single object and invoke passive external knowledge sources called \emph{schemata} that prescribe state updates. To use a video game as an illustration, two enemies of the same type will share schemata but will have separate object files to encode their distinct state (e.g., health, position). We propose to use attention to determine which object files to update, the selection of schemata, and the propagation of information between object files. The resulting architecture is a drop-in replacement conforming to the same input-output interface as normal recurrent networks (e.g., LSTM, GRU) yet achieves substantially better generalization on environments that have multiple object tokens of the same type, including a challenging intuitive physics benchmark.
 
\end{abstract}

\section{Introduction}

An intelligent agent that interacts with its world must not only perceive objects but must also remember its past experience with these objects. The wicker chair in one's living room is not just a chair, it is the chair which has an unsteady leg and easily tips. Your keys may not be visible, but you recall placing them on the ledge by the door. The annoying fly buzzing in your left ear is the same fly you saw earlier which landed on the table. 

Visual cognition requires a short-term memory that keeps track of an object's location, properties, and history. In the cognitive science literature, this particular form of state memory is often referred to as an \emph{object file} \citep{Kahneman1992}, which we'll abbreviate as \OF. An \OF\ serves as a temporally persistent reference to an external object, permitting object constancy and permanence as the object and the viewer move in the world. 

Complementary to information in the \OF\ is abstract knowledge about the dynamics and behavior of an object. We refer to this latter type of knowledge as a \emph{schema} (plural \emph{schemata}), another term borrowed from the cognitive-science literature. The combination of \OFs\ and schemata is sufficient to predict future states of object-structured environments, critical for planning and goal-seeking behavior. 

To model a complex, structured visual environment, multiple \OFs\ must be maintained in parallel. Consider scenes like a PacMan video-game screen in which the ghosts chase the PacMan, a public square or sports field in which people interact with one another, or a pool table with rolling and colliding balls. In each of these environments, multiple instances of the same object class are present; all operate according to fundamentally similar dynamics. To ensure systematic modeling of the environment, the same dynamics must be applied to multiple object instances.
Toward this goal, we propose a method of separately representing the state of an individual object---via an \OF---and how its state evolves over time---via a schema.


\renewcommand{\arraystretch}{.85} 
\begin{SCfigure}[.6][tb]
\centering
\begin{minipage}{0.18\linewidth}
    \includegraphics[height=1.2in,width=1in]{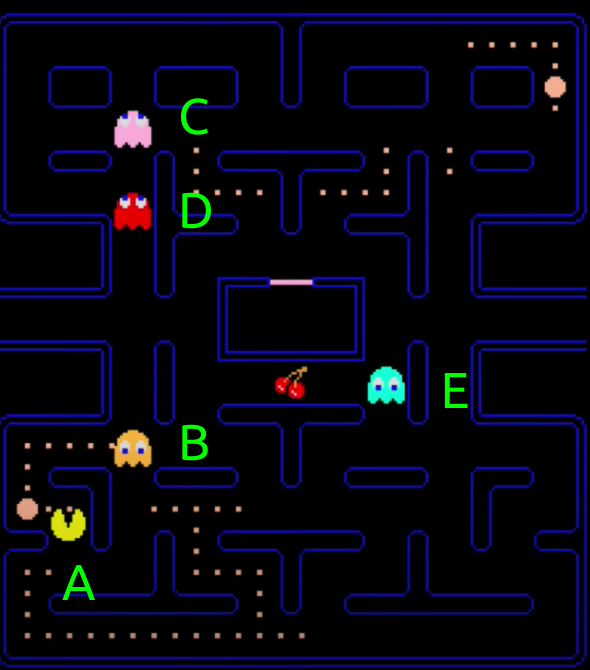}
\end{minipage}
\begin{minipage}{0.18\linewidth}
    \includegraphics[height=1.2in,width=1in]{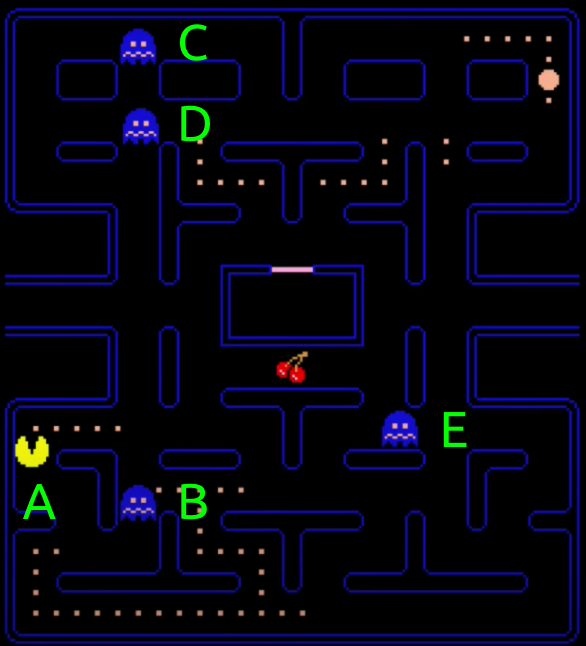}
\end{minipage}
\begin{minipage}{0.30\linewidth}
\scalebox{0.8}{
\begin{tabular}{|c|c|c|c|}
\hline
\tiny Object & \tiny Schema 1 & \tiny Schema 2 & \tiny Schema 3 \\
\tiny files & \tiny PacMan & \tiny Normal & \tiny Scared \\ 
 &  \tiny (\OFs) & \tiny Ghost & \tiny Ghost \\ \hline
 \multicolumn{4}{|c|}{\tiny Left Frame} \\ \hline
\tiny A & \cellcolor{blue!15}\tiny\checkmark &   &    \\ \hline
\tiny B &  &  \cellcolor{blue!15}\tiny\checkmark &    \\ \hline
\tiny C &   & \cellcolor{blue!15}\tiny\checkmark &    \\ \hline
\tiny D &   &  \cellcolor{blue!15}\tiny\checkmark &    \\ \hline
\tiny E &  &  \cellcolor{blue!15}\tiny\checkmark  & \\ \hline
\multicolumn{4}{|c|}{\tiny Right Frame} \\ \hline
\tiny A & \cellcolor{blue!15}\tiny\checkmark &  &    \\ \hline
\tiny B &  &   & \cellcolor{blue!15}\tiny\checkmark   \\ \hline
\tiny C &   &  & \cellcolor{blue!15}\tiny\checkmark   \\ \hline
\tiny D &   &  & \cellcolor{blue!15}\tiny\checkmark   \\ \hline
\tiny E &  &    & \cellcolor{blue!15}\tiny\checkmark \\ \hline
\end{tabular}
}
\end{minipage}
\caption{Two successive frames of PacMan, illustrating the factorization of knowledge. Each ghost is 
represented by a persistent \OF\ (maintaining its location and velocity), but all ghosts operate according to one of two schemata, depending on whether the ghost is in a \textit{normal} or \textit{scared} state.}
\label{fig:examplegame}
\end{SCfigure}

Object-oriented programming (OOP) provides a metaphor for thinking about 
the relationship between \OFs\ and schemata. In OOP, each \emph{object} is 
an instantiation of an object class and it has a self-contained collection 
of variables whose values are specific to that object 
and \textit{methods} that operate on all instances of the same class.
The relation between objects and methods mirrors the relationship between our
\OFs\ and schemata.
In both OOP and our view of visual cognition, a key principle is the
\emph{encapsulation} of knowledge: internal details of objects (\OFs)
are hidden from other objects (\OFs), and methods (schemata) are accessible 
to all and only objects (\OFs) to which they are applicable.

The modularity of knowledge in OOP supports
human programmers in writing code that is readily debugged, extended, and reused.
We conjecture that the corresponding modularity of \OFs\ and 
schemata will lead to neural-network models with more efficient learning and
more robust generalization, thanks to appropriate disentangling
and separation of concerns.

\begin{SCfigure}[.6][b!]
    \centering
    \includegraphics[width=3.5in]{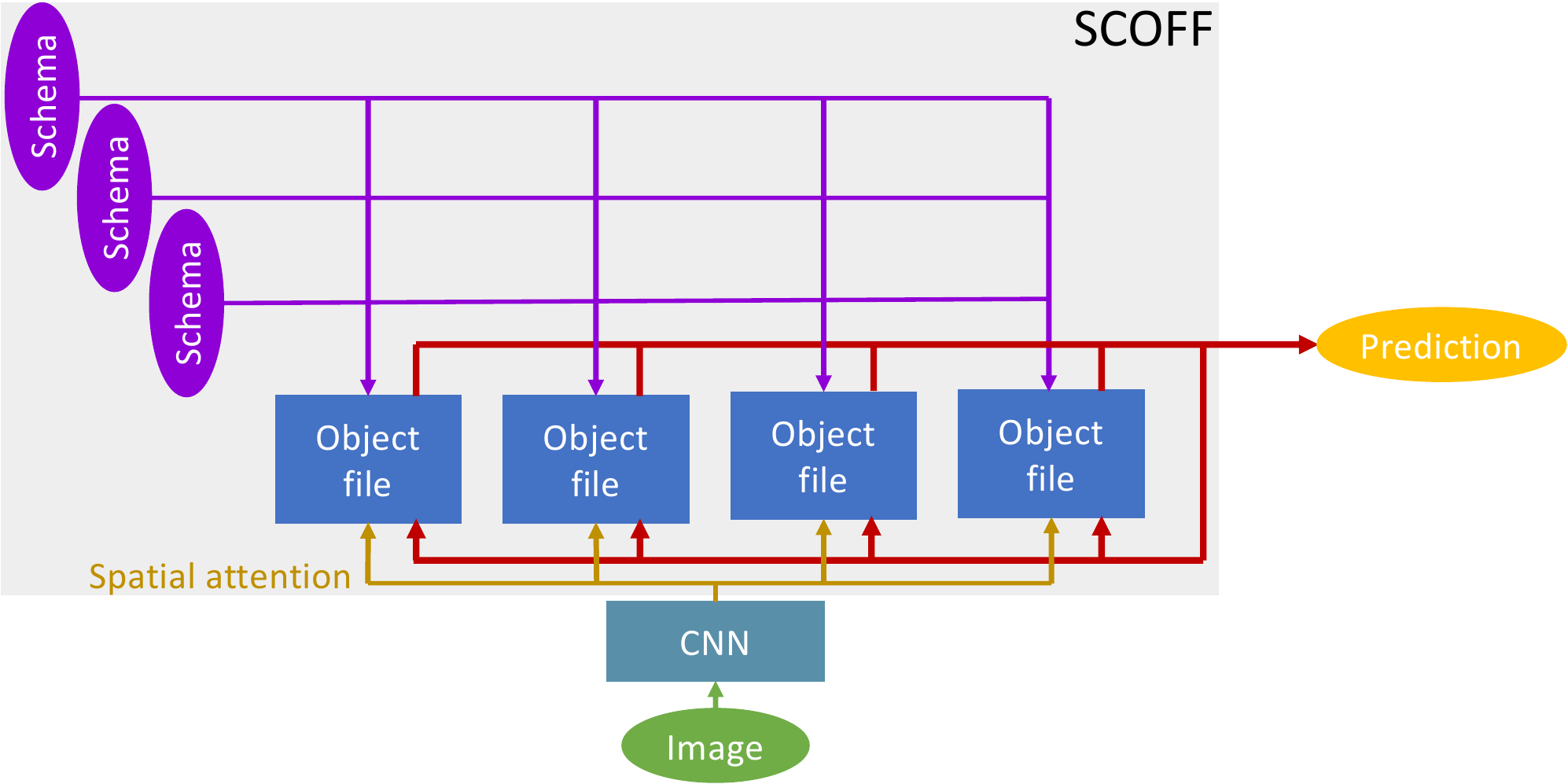}
    \caption{\emph{Proposed \modelname\ model.}
    Schemata are sets of parameters that specify the dynamics of objects.
    Object files (\OFs) are active modules that maintain the time-varying state of an object, seek information from the input, and select schemata for updating, and transmit information to 
    other object files. Through spatial attention, \OFs\ 
    compete and select different regions of the input.
    \label{fig:model}}
\end{SCfigure}

Modularity is the guiding principle of the model we propose, which we call \modelname, an acronym
for \emph{\underline{sc}hema / \underline{o}bject-\underline{f}ile \underline{f}actorization}.
Like other neural net models with external memory \citep[e.g.,][]{das1993,graves2016hybrid,Sukhbaatar2015}, 
\modelname\ includes
a set of slots which are each designed to contain an \OF\ 
(Figure~\ref{fig:model}).
In contrast to most previous external memory models, the slots are
not passive contents waiting to be read or written by an active
process, but are dynamic, modular elements that seek information 
in the environment that is relevant to the object they represent,
and when critical information is observed, they update their states,
possibly via information provided by other \OFs. Event-based OOP is 
a good metaphor for this active process, where external events can 
trigger the action of objects.

As Figure~\ref{fig:model} suggests, there is a factorization of
\emph{declarative} knowledge---the location, properties, and history of an object, as contained in the \OFs---and \emph{procedural} knowledge---the rules of object behavior, as contained in the schemata.
Whereas declarative knowledge can change rapidly, procedural knowledge is 
more stable over time. This factorization
allows any schema to be applied to any \OF\ as deemed appropriate.
The model design ensures \emph{systematicity} in the operation of a 
schema, regardless of the slot to which an \OF\ is assigned. Similarly, an
\OF\  can access any applicable schema regardless of which slot it sits in.
Furthermore, a schema can be applied to multiple \OFs\ at once, and 
multiple schemata could be applied to an \OF\ (e.g., Figure~\ref{fig:examplegame}). 
In OOP, systematicity is similarly achieved  by virtue of the fact that the same method can be applied to any object instantiation and that multiple methods exist which can be applied to an object of the appropriate type.


Our key contribution is to demonstrate the feasibility and benefit of factorizing declarative knowledge (the location, properties, and history of 
an object)  and procedural knowledge (the way objects behave). 
This factorization enforces not only an important
form of systematicity, but also of exchangeability: the model behaves
exactly the same regardless of the assignment of schemata to schemata-slots
and the assignment of objects to \OF-slots, i.e., the
neural network operates on a {\em set} of objects
and a {\em set} of schemata. With this factorization,
we find improved accuracy of next-state prediction models and improved
interpretability of learned parameters.

\section{The schemata / object-file factorization (\modelname) model}

\modelname, shown in 
Figure~\ref{fig:model}, is an architectural backbone that supports 
the separation of procedural and declarative knowledge about 
dynamical entities (objects) in an input sequence. The input sequence
$\{ \vx_1, \ldots, \vx_t, \ldots,  \vx_T\}$, indexed by time step $t$ is
processed by a neural encoder (e.g., a fully convolutional net for
images) to obtain a deep embedding, $\{ \vz_1, \ldots, \vz_t, \ldots, \vz_T\}$, 
which then drives a network with $n_f$ \OFs\ and  $n_s$ schemata.

\OFs\ are active processing components that maintain and update their
internal state.  Essentially, an \OF\ is a layer of GRU \citep{chung2014empirical} or LSTM \citep{hochreiter1997long} units with three additional 
bits of machinery, which we now describe.
\begin{enumerate}
\item
Our earlier metaphor identifying \OFs\ in \modelname\ with objects in OOP is apropos 
in the sense that \OFs\ are \emph{event driven}. \OFs\ operate in a temporal loop,
continuously awaiting relevant input signals. Relevance is determined by an attention-based soft
competition among \OFs:  the current input serves a key that is matched to a query generated 
from the state of each \OF; based on the goodness of match, each \OF\ is provided with a value 
extracted from the input.

\item
Each \OF\ performs a one-step update of its state layer of GRU or LSTM units,
conditioned on the input signal received. The weight parameters needed to perform this update,
which we will denote generically as $\vtheta$, are not---as in a standard GRU or LSTM---internal to
the layer but rather are provided externally. Each schema $j$ is nothing more than a set of parameters $\vtheta_j$ which can be plugged into this layer. \modelname\ uses a key-value attention
mechanism to perform Gumbel-based hard selection of the appropriate schema (parameters).
%
\item
\OFs\ may pass information to other \OFs\ (analogous to arguments being passed to a method in OOP),
again using a soft attention mechanism by which each \OF\ queries all other \OFs.
Keys provided by the other \OFs\ are matched to the query, and a soft selection of the best matching \OFs\ determines 
the weighting  of values transmitted by the other \OFs.

\end{enumerate}
This operation cycle ensures that \OFs\ can update their state in response to both the external
input and the internal state comprised
of all the \OFs' contents. This updating is an extra wrapper around the ordinary update that
takes place in a GRU or LSTM layer. It provides additional flexibility in that 
(1) The external input is routed to \OFs\ contingent on their internal state,
(2) \OFs\ can switch their dynamics from one time step to the next conditioned on their internal state, 
(3) \OFs\ are modular in that they do not communicate with one another except via state-dependent selective message 
passing.

\OFs\ are placeholders in that a particular \OF\ has no weight parameter specific to that \OF.
Instead, the parameters are provided from two sources: either the schemata or a pool of 
generic parameters shared by the $n_f$ \OFs. This generic parameter pool is used to implement 
key-value attention over the input, the schemata, and communication among \OFs. The sharing
of schemata ensures systematicity; the sharing of the generic parameter pool ensures
\emph{exchangeability}---model behavior is unaffected by the assignment of object
instances to \OF\ slots.

\subsection{\modelname\ specifics}

\begin{algorithm}[bt]
    \caption{ \modelname\ model}
   \label{alg:attention}
   \begin{algorithmic}
   \footnotesize
    \STATE{\bfseries \itshape Input:} Current sequence element, $\vx_t$ and
    previous \OF\ state, $\{ \vh_{t-1,k} | ~k \in \{1, \ldots, n_f \} \}$

   \item[]
   \STATE{\bfseries \itshape Step 1: Process image by position $p$ with fully convolutional net}
  \STATE \bull $\vc_{p} = [ \mathrm{CNN}  (\vx_{t})]_p$
  \STATE \bull $\va_{p} = [\vc_{p} ~\ve_{p}]$
  ~~~\textit{(concatenate encoding of position to CNN output)}

  \item[]
   \STATE{\bfseries \itshape Step 2: Soft competition among \OFs\ to select regions of the input to process}
  \STATE \bull $\vq_k = \vW^q \vh_{t-1,k}$
  \STATE \bull $s_{k,p} = \mathrm{softmax}_{k}\big(\frac{\vq_k \vkappa_{k,p}}{\sqrt{d_e}}\big),  \text{where } \vkappa_{k,p} = (\va_{p}\vW^e)^\mathrm{T}$ 
  \STATE \bull $\vz_{k} = \sum_{p} s_{k,p} \vv_{p} \quad \text{where } \vv_{p} = \va_{p} \vW^v \quad \forall ~k  \in \{1, \ldots, n_f\}$
   

   


   \item[]
   
   \STATE{\bfseries \itshape Step 3: \OFs\ pick the most relevant schema and update}
   \STATE \bull $\widetilde{\vh}_{t,k,j} = \mathrm{GRU}_{\vtheta_j}\left( \vz_k, \vh_{t-1,k} \right) \quad \forall k \in \{1, \ldots, n_f\}, j \in \{1,\ldots, n_s\}$
    
   \STATE \bull $\widetilde{\vq}_k = \vh_{t-1,k} \widetilde{\vW}^q$
   
   \STATE \bull $i_k = \mathrm{argmax}_j  \left(
   \widetilde{\vq}_k \widetilde{\vkappa}_{k,j} 
   + \gamma \right), \text{ where }
   \widetilde{\vkappa}_{k,j} = (\widetilde{\vh}_{t,k,j} \widetilde{\vW}^e )^\mathrm{T} \text{ and } \gamma \sim \mathrm{Gumbel} (0,1)$
   
   \STATE \bull $\vh_{t,k}=\widetilde{\vh}_{t, k, i_k}$
       
   \item[]
   \STATE{\bfseries \itshape Step 4:  Soft competition among \OFs\ to transmit relevant information to each \OF} 
   
   \STATE \bull $\widehat{\vq}_k=\vh_{t-1,k} \widehat{\vW}^{q} \quad \forall k \{1, \ldots, n_f\}  $ 
              
   \STATE \bull $s_{k,k'} = \mathrm{softmax}_{k'} \left( \frac{\widehat{\vq}_{k} \widehat{\vkappa}_{k'}
   }{\sqrt{d_e}}\right) \text{ where }
   \widehat{\vkappa}_{k'} = ( \vh_{t,k'} \widehat{\vW}^{e}  )^\mathrm{T} \quad \forall~ k,  ~k' \in \{1, \ldots, n_f\}
   $
   
   \STATE \bull $\vh_{t,k} \gets \vh_{t,k} + 
              \sum_{k'} s_{k,k'}
              \widehat{\vv}_{k'}
              \text{~~where } \widehat{\vv}_{k'} = \vh_{t,k'} \widehat{\vW}^{v}
              \quad \forall k \in \{1, \ldots, n_f\} 
   $

\end{algorithmic}
\end{algorithm}



Algorithm~\ref{alg:attention} provides a precise specification of \modelname\  broken into four steps. 
Step 1 is external to \modelname\ and involves processing an image input to obtain a deep embedding. The processing is performed by a fully Convolutional Neural Net (CNN) that preserves positional information
(typically $64\times64$ in our simulations),
and for each position $p$ encodes the processed input $\vc_p$ and concatenates
a learned encoding of position, $\ve_p$ to produce a position-specific
hidden state.
(\modelname\ can also handle non-visual inputs, such as vectors or discrete
tokens; in this case the CNN is replaced by an MLP.) The subsequent core steps are as
follows.




{\bfseries \itshape Step 2: Soft competition among \OFs\ to select regions of the input to process.} 
The state of each \OF\ $k$, $\vh_{t-1,k}$, is used to form a query,
$\vq_k$, which determines the input positions that it will attend to.
The query is matched to a set of position-specific input keys, 
$\vkappa_{k,p}$ for position $p$, producing a position-specific match score, $s_{k,p}$. Soft position-specific competition among the \OFs\ results in the \OFs\ selecting complementary image regions to attend to. The contents of the attended positions are combined yielding an
\OF-specific input encoding, $\vz_k$.


{\bfseries \itshape Step 3:  \OFs\ pick the most relevant schema and update.}
\OF\ $k$ picks one schema via attention as follows. \OF\ $k$ binds to \emph{each} schema $j$, and then performs a hypothetical update, yielding $\widetilde{\vh}_{t,k,j}$. In experiments below, the \OF\ state is maintained by a GRU layer, and schema $j$ is a parameterization of the GRU,  denoted $\smash{\mathrm{GRU}_{ \vtheta_j}}$, which determines the update. The previous state of the  \OF, $\vh_{t-1,k}$ serves as a query in key-value attention against a key derived from the hypothetical updated state, $\widetilde{\vh}_{t,k,j}$. The schema $i_k$ corresponding to the best query-key match for \OF\ $k$ is used to update \OF\ $k$'s state.  
Selection is based on the straight-through Gumbel-softmax method \citep{jang2016categorical}, which makes a hard choice
during forward propagation and during backward propagation, it considers a softened version of the output to permit gradient propagation to non-selected schemata.

{\bfseries \itshape Step 4:  Soft competition among \OFs\ to transmit relevant information to each \OF.} 
This step allows for interactions among \OFs.
Each  \OF\ $k$ queries other \OFs\ for information relevant to its update as follows.
The \OF's previous state, $\vh_{t-1,k}$, is used to form a query, $\widehat{\vq}_k$, in an
attention mechanism against a key derived from the new state of each other \OF\ $k'$,  
$\widehat{\vkappa}_{k'}$, and softmax selection ($s_{k,k'}$) is used to obtain weighted information from other \OFs, $\widehat{\vv}_k'$, into \OF\ $k$'s state.


\textit{Number of Parameters.}  \modelname\ can be used as a drop-in replacement for a  LSTM/GRU layer. There is a subtlety that must be considered for successful integration. If the total size of the hidden state is kept the same, integrating \modelname\ dramatically reduces the total number of recurrent parameters in the model because of its block-sparse structure. The majority of \modelname\ parameters are the schemata, $\{ \vtheta_j | j\in\{1, \ldots, n_s\} \}$. The remaining parameters are those of query
($\vW^q$, $\widetilde{\vW}^q$,  $\widehat{\vW}^q$), key
($\vW^e$, $\widetilde{\vW}^e$,  $\widehat{\vW}^e$), and value
$\vW^v$, $\widehat{\vW}^v$) functions. Note that these linear functions could be replaced by nonlinear functions. Its also interesting to note that \modelname\ actually has far fewer parameters than other modular architectures (e.g., RIMs, Recurrent Entity Networks, Neural Module Networks) when $n_s < n_f$ (which holds true for most of our experiments), because of the potential one-to-many mapping between schemata and \OFs. Optionally, at {\itshape Step 2}, during training instead of activating all the \OFs,  we can use a sparse attention, to only activate a subset of \OFs\ that are relevant at that time step $t$, and only update the state of the activated \OFs.  We note that this is not specific to the proposed method, but the performance of the method can be improved by only selectively activating the relevant \OFs. 




\vspace{-2mm}
\section{Related Work}
\vspace{-2mm}

\textbf{CNNs}. \modelname\ applies the same knowledge (schemata) to multiple 
\OFs, yielding systematicity. Similarly, 
a CNN is a highly restrictive instantiation of this same notion, where 
knowledge---in the form of a convolutional filter---is applied uniformly to every location in an image,
yielding equivariance. \modelname\ is a more flexible architecture than a CNN in that \OFs\ 
are defined by abstract notion of objects not a physical patch of an image, and schemata are applied
dynamically and flexibly over time, not in a fixed, rigid manner as the filters in a CNN.

\textbf{Memory Networks}. A variety of existing models leverage an external slot-based,
content-addressible memory \citep[e.g.,][]{graves2016hybrid,Sukhbaatar2015}. The memory
slots are passive elements that are processed by a differentiable neural controller such as an
LSTM. Whereas traditional memory networks have many dumb memory cells and one smart controller,
\modelname\ has many smart memory cells---the \OFs---which also act as local controllers.
However, \modelname\ shares the notion with memory networks that the same knowledge is applied
systematically to every cell.
 


\textbf{Relational RNN (RMC)}. The RMC \citep{santoro2018relational} has a multi-head attention
mechanism which allows it to share information between multiple memory locations.
RMC is like Memory Networks in that dynamics are driven by a central controller: memory is
used to condition the dynamics of an RNN.

\textbf{Recurrent Entity Networks}.  
\citet{henaff2016tracking} describe a collection of recurrent modules that update independently 
and in parallel in response to each input in a sequence. The module outputs are integrated to form 
a response. It shares a modular architecture with \modelname, but the modules have a fixed function 
and do not directly communicate with one another. This earlier work focused on language not images.

\textbf{Recurrent Independent Mechanisms (RIMs)}.
RIMs \citep{goyal2019recurrent} are a key inspiration for our work.  RIMs are a modular neural
architecture consisting of an ensemble of dynamical components which have sparse interactions through 
the bottleneck of attention. Each RIM module has at its core an LSTM layer 
\citep{hochreiter1997long}. A RIM module is much like our \OF.  Both are meant to be
dynamical entities with a time-evolving state. However, in contrast to \OFs\ in \modelname, RIM 
modules operate according to fixed dynamics and each RIM module is specialized for a particular
computation. RIM modules are thus not interchangeable.


\textbf{Neural Module Networks}. A modular module network \citep{jacobs1991adaptive, bottou1991framework,ronco1996modular, reed2015neural, andreas2016neural,rosenbaum2017routing, fernando2017pathnet, shazeer2017outrageously, kirsch2018modular, rosenbaum2019routing}  has an architecture which is composed 
dynamically from several neural modules, where each module is meant to perform a distinct function. Each module has a fixed function, unlike our \OFs, and modules are applied one at a time, in contrast
to our \OFs, which can update and be used for prediction in parallel.



\vspace{-2mm}
\section{Methodology}
\vspace{-2mm}
\modelname\ is a drop-in replacement for a standard LSTM or GRU layer, conforming to the 
same input-output interface.
Because of their interchangeability, we compare \modelname\ to LSTM and GRUs. We also compare \modelname\ to 
two  alternative modular architectures: \textit{RMC}, a memory based relational recurrent model with attention between memory elements and hidden states  \citep{santoro2018relational}, and \textit{Recurrent Independent Mechanisms (RIMs)}, a modular memory based 
on a single layered recurrent model with attention modulated input and communication between 
modules \citep{goyal2019recurrent}. 

In all simulations we present, the input is a video sequence, each frame of which is preprocessed by a CNN backbone.
We consider two task types: video prediction and reinforcement learning (RL). For video prediction, the simulation output
is a prediction of the next frame in the sequence. For RL, the output is an action choice.
In both cases, \modelname's internal state is mapped to an output in the following manner. The state of each \OF\ is
remapped by a siamese network that transforms the state. (By siamese network, we mean that every \OF\ is remapped by
the same function.) The transformed state is concatenated and used by an attention-based mechanism operating over the
\OF\ slots to ensure exchangeability of the \OFs. In the case of video prediction, a deconvolutional backbone yields 
an image as output;  in the case of reinforcement learning, the output is a distribution over actions.

The heart of the model is a single \modelname\ layer, consisting of $n_f$ \OFs\ and $n_s$ schemata.
Most simulation details are contained in section \ref{sec:hyperparameters} of the Appendix, but we summarize some key points
here. Unless otherwise indicated, we always train in an end-to-end fashing using the 
Adam~\citep{Kingma2014} optimizer with a learning rate of 0.0001 and momentum of 0.9. As a default, we use
$n_f=6$ and $n_s=4$, except where we are specifically exploring the effects of manipulating these hyperparameters.
We include more experimental results  in the Appendix, and we will release the code.

\vspace{-2mm}
\section{Experiments}
\vspace{-2mm}

Our experiments address the following questions.
(1) Does \modelname\ successfully factorize knowledge into \OFs\ and schemata? 
(2) Do the learned schemata have semantically meaningful interpretations? 
(3) Is the factorization of knowledge into object files and schemata helpful in downstream tasks?
(4) Does \modelname\ outperform state-of-the-art approaches, both modular and non-modular, which lack \modelname's strong inductive bias toward systematicity and knowledge factorization?


\begin{figure}[b!]
    \includegraphics[width=0.53\textwidth]{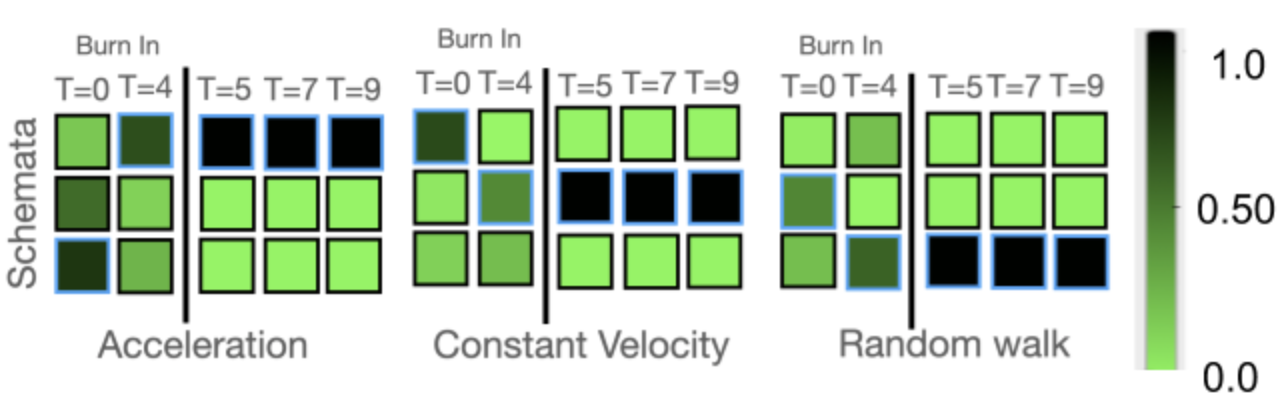}
    \includegraphics[width=0.46\textwidth]{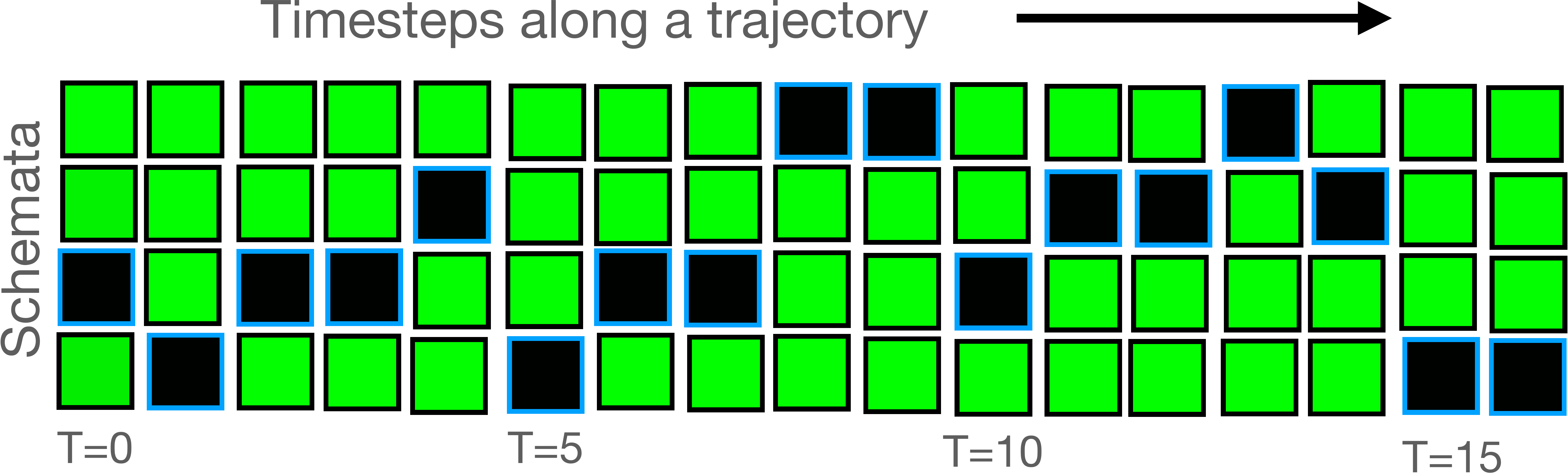} 
    \caption{[\emph{left panel}] Single object sequences with three possible dynamics of motion. After about five video frames (burn-in period), \modelname\ locks into the type of motion and activates a corresponding schema to predict future states. Relative activation of the three schemata indicated by the color bar. The selected schema indicated by the faint blue border.
    [\emph{right panel}]
     \textit{RL Maze task.} The agent learns to navigate a maze of randomly interconnected rooms to find a key. Array indicates schema activation over a sequence of steps along a particular trajectory.}
     \label{fig:simple_working_experiment}
\end{figure}

We begin with a series of experiments involving greyscale synthetic video sequences consisting of a single ball-like object  moving over time according to switchable dynamics. We model this scenario with \modelname\ having a single ($n_f = 1$) \OF\ and the number of schemata matching the number of different types of dynamics.

{\bfseries \itshape Single object with fixed dynamics.} To demonstrate that the proposed model is able to 
factorize knowledge,  we consider video scenes in which a single object, starting in a random location, has dynamics that cause it to either
(a) accelerate in a particular direction, (b) move at a constant velocity in a particular direction, or (c) take a random
walk with a constant velocity. Following training, \modelname\ can predict trajectories after being shown the first
few frames. It does so by activating a schema that has a one-to-one correspondence with the three types of dynamics 
(Figure 3, left panel), leading to \emph{interpretable semantics} and a clean factorization of knowledge.
For additional details, refer to Appendix~\ref{sec:bouncing_ball}. 


\begin{SCfigure}[5]
    \centering
    \includegraphics[width=0.6\textwidth]{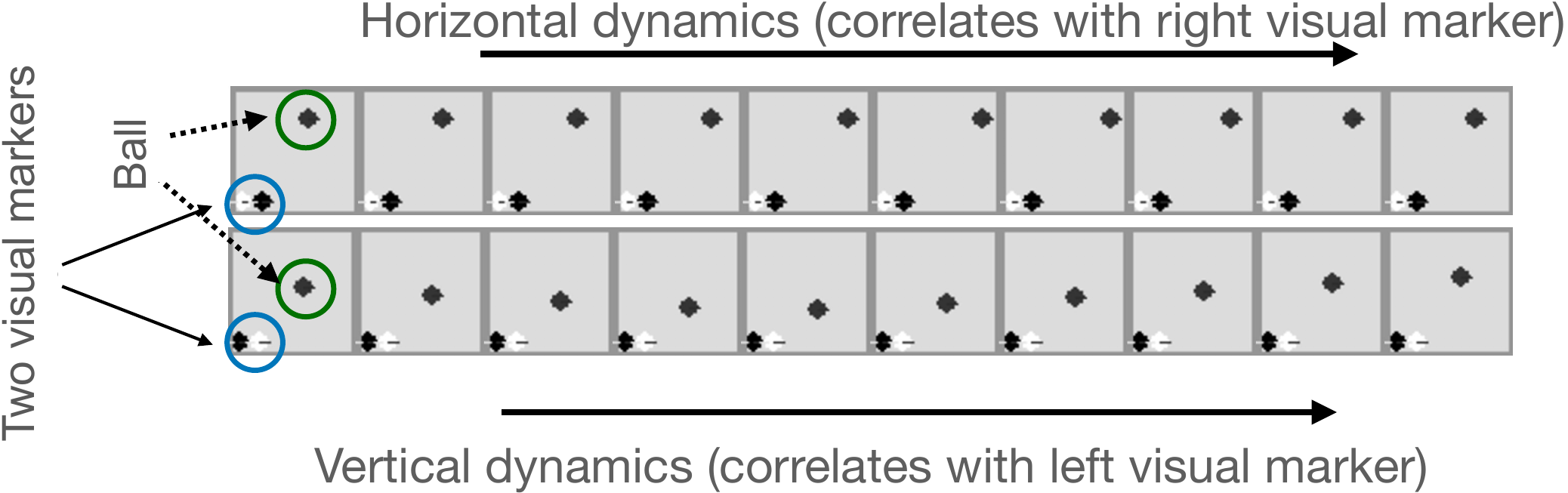}
    \caption{\textit{Switching dynamics task.} Sequence of steps showing horizontal (top row) and vertical (bottom row) dynamics of a ball. Frames contain visual markers that indicate the current dynamics.}
    \label{fig:dynamics_visual}
\end{SCfigure}

{\bfseries \itshape Single object with switching dynamics.}
In this experiment, video scenes contain a single object, starting in a random location, operating according two one of two dynamics: vertical and horizontal motion. In contrast to the previous experiment, the object can switch dynamics in the course of a sequence. The current dynamics are indicated by markers in the image (see Figure~\ref{fig:dynamics_visual}) which \modelname\ must learn to detect in order to select the corresponding schema.  In this experiment, we find that \modelname\ is able to learn a one-to-one correspondence with two types of dynamics, yielding high-precision prediction of the next frame, even on the first frame after a switch.

{\bfseries \itshape Single object with switching dynamics in an RL paradigm.}
In this experiment, we use the partially-observable GotoObjMaze environment from \cite{chevalier2018babyai}, in which an agent must learn to navigate to navigate a 2D multi-room grid world to locate distinct objects such as a key. 
The world consists of $6\times6$
square rooms that are randomly connected through doors to form a $3 \times 3$ maze of rooms. The agent's view of the
world is an egocentric $5\times5$ patch. Our experiment involves
$n_s=4$ schemata whose activation pattern over time steps is shown in the right panel of Figure~\ref{fig:dynamics_visual}.
The schema activation pattern is interpretable; for example, schema 4 is triggered when the `key' is in the agent's
field of view (see Appendix, Figure~\ref{fig:babyai} and Section~\ref{sec:reinforcement_learning}).
To obtain quantitative evaluations, we test transfer by increasing the room size to $10\times10$ during testing.
\modelname\ is able to successfully reach its goal on 82\% of trials, whereas a GRU baseline succeeds only 56\% of
trials. Even an overparameterized GRU, which matches \modelname\ in number of free parameters, still succeeds
less often---on 74\% of trials. All models are trained for an
equal number of time steps. To summarize, the factorization of knowledge in \modelname\ leads not only to better
next-step prediction but also improve performance on a downstream control task.

\begin{figure}[bt]
    \centering
    \subfigure[\textbf{4Balls}]{\includegraphics[ width=0.5\textwidth, trim={2.5cm 1.3cm 3cm 1cm},clip]{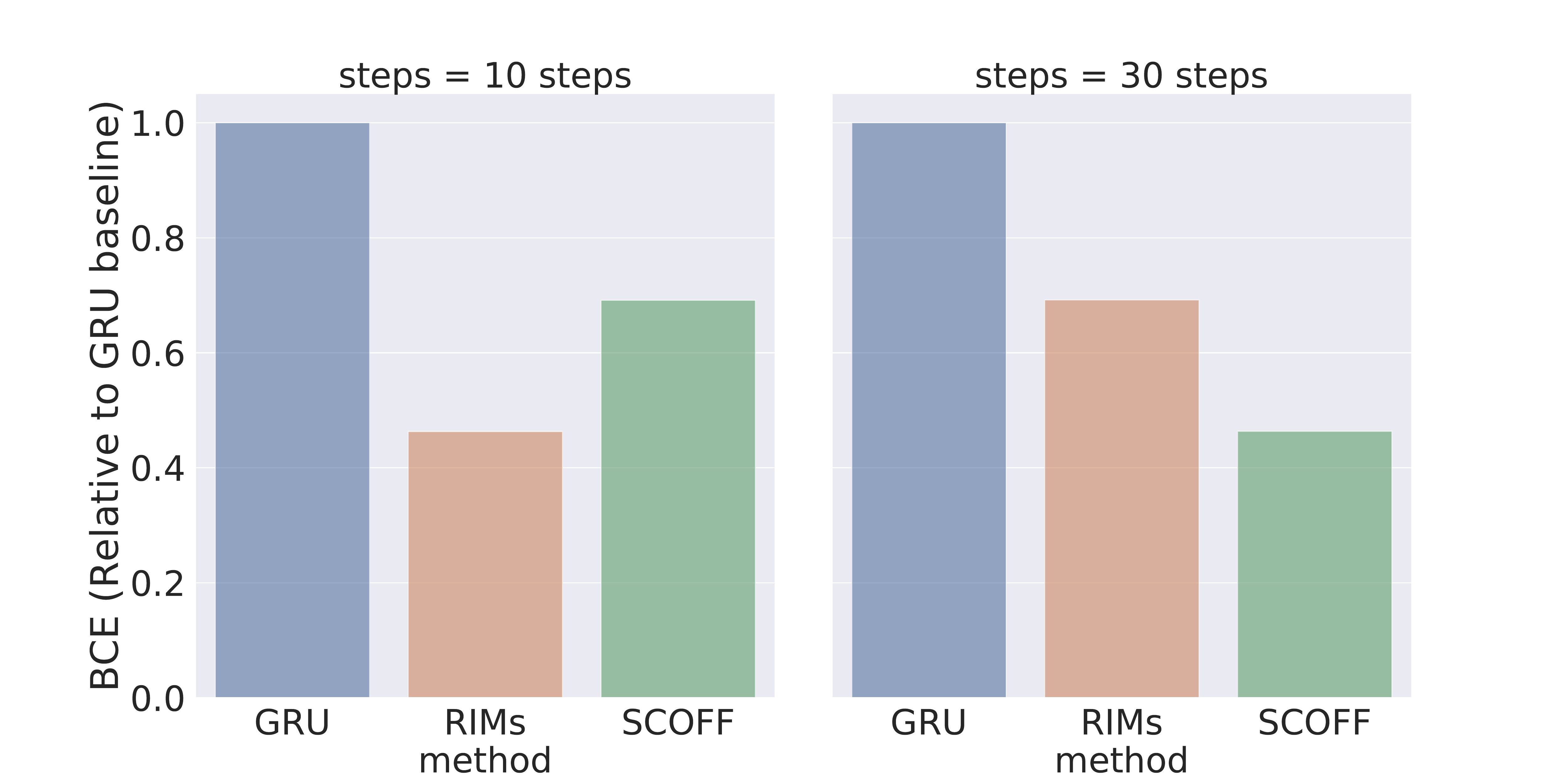}} \hspace{-1em}
    \subfigure[\textbf{Curtain}]{\includegraphics[ width=0.5\textwidth, trim={2.5cm 1.3cm 3cm 1cm},clip]{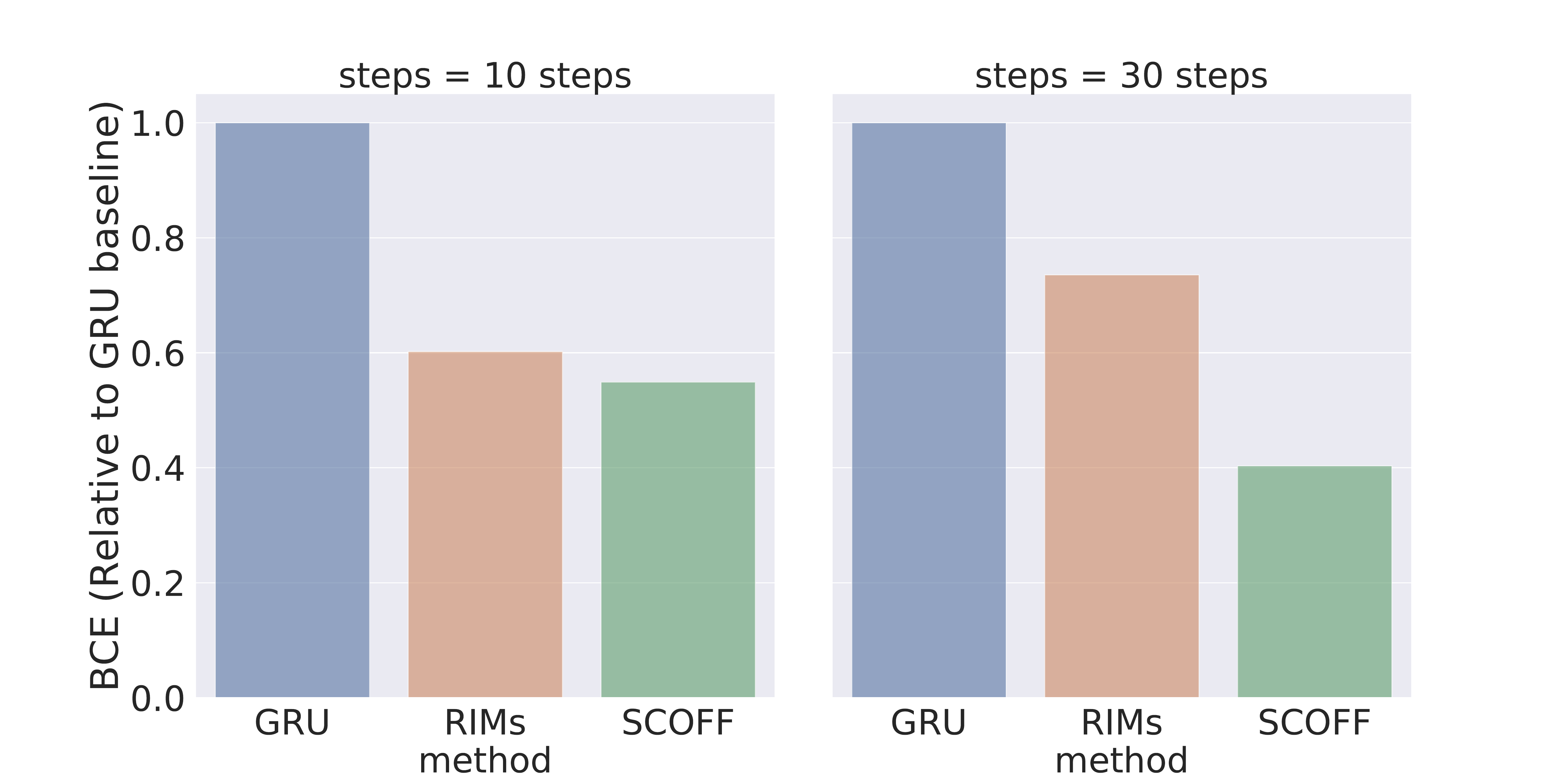}} 
    \subfigure[\textbf{678Balls}]{\includegraphics[ width=0.49\textwidth, trim={2.5cm 1.3cm 3cm 1cm},clip]{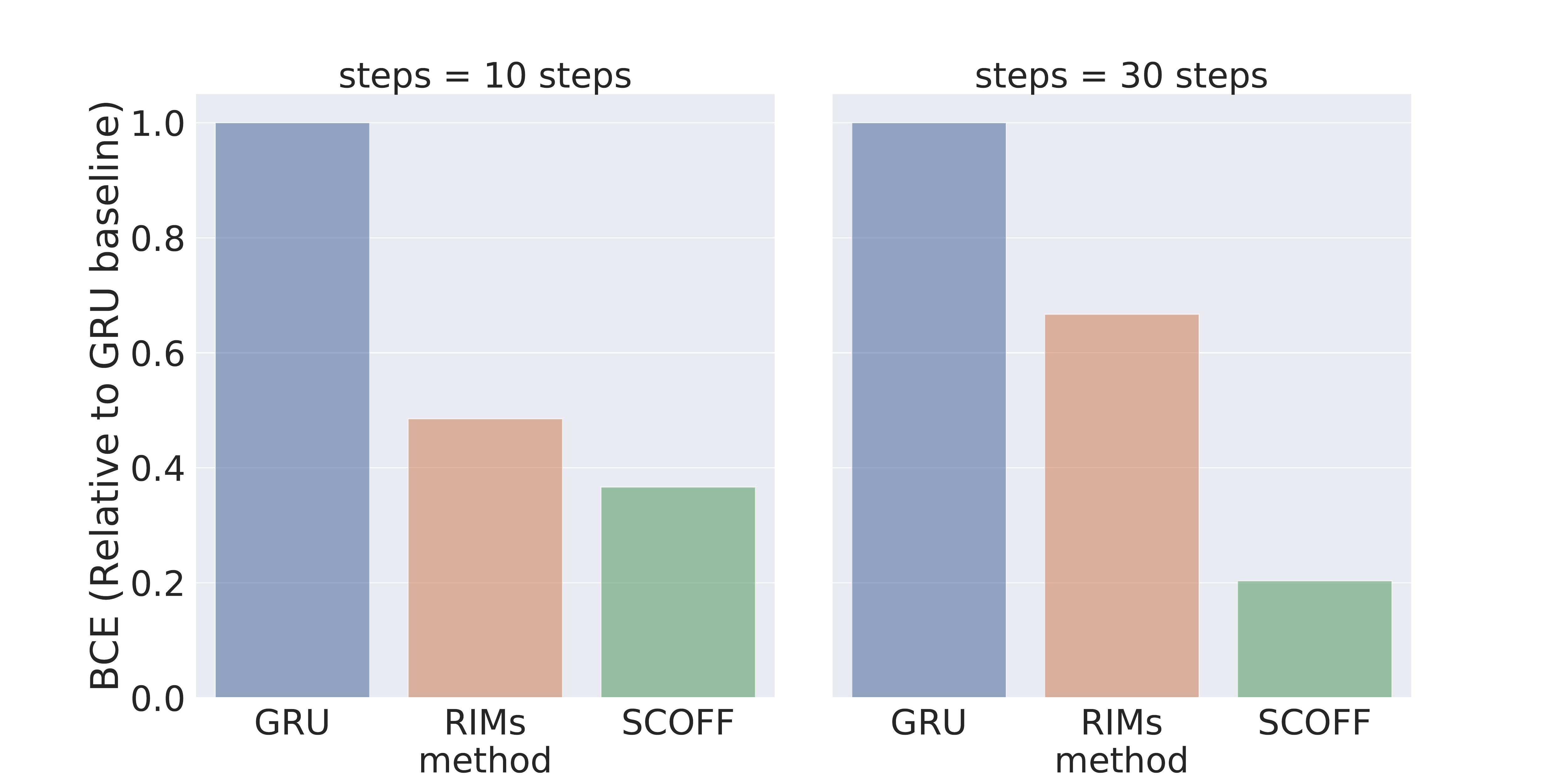}}
    \subfigure[\textbf{Coloured678Balls}]{\includegraphics[ width=0.5\textwidth,trim={2.5cm 1.3cm 3cm 1cm},clip]{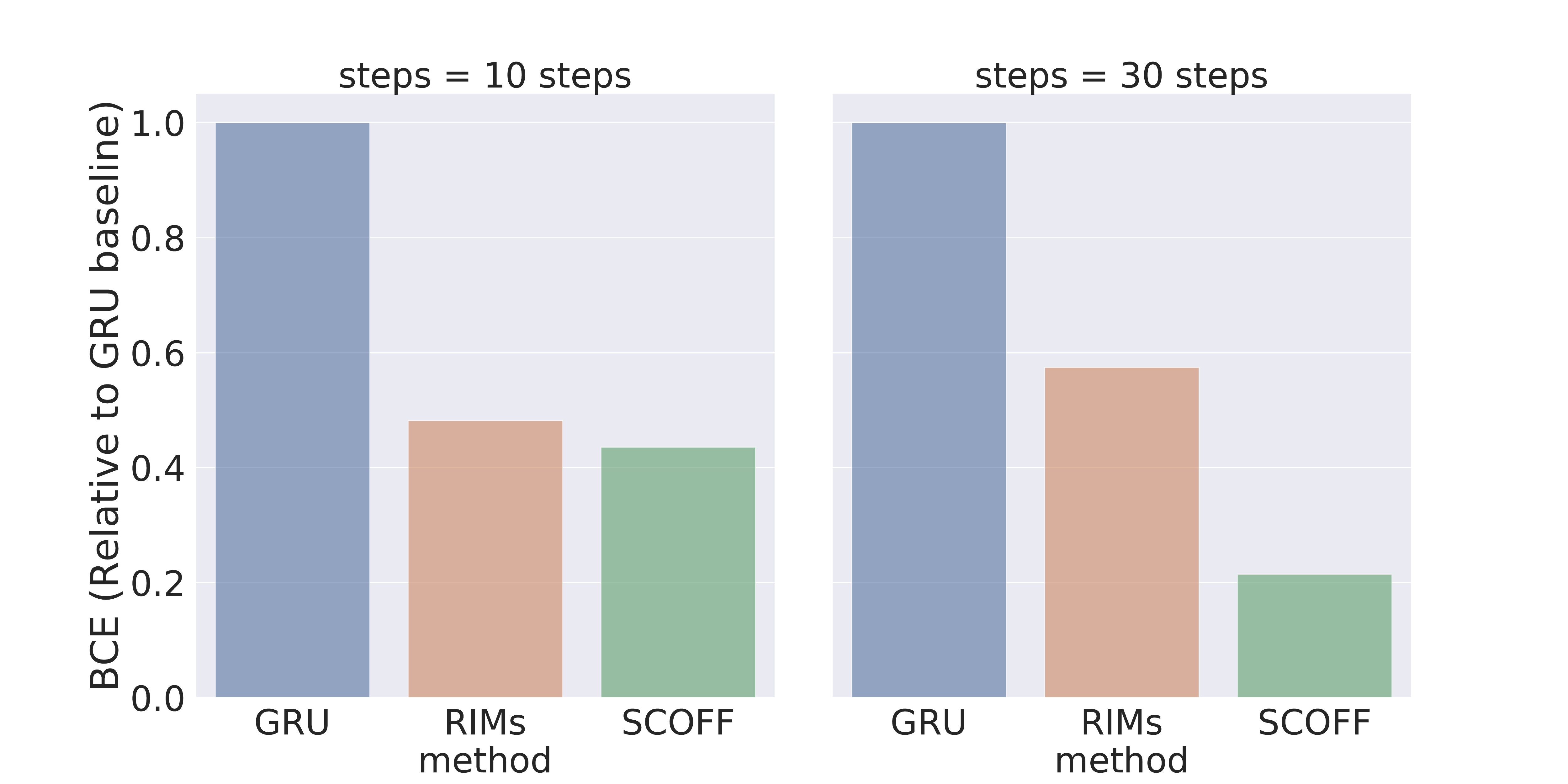}}\par

    \caption{\textit{Bouncing ball motion.} Error relative to GRU baseline on 10- and 30-frame video prediction
    of multiple-object video sequences  Predictions are based on 15 frames of ground truth. The advantage of
    \modelname\ is amplified as the number of balls increases (4Balls versus 678Balls) and as predictions require
    greater look ahead (10 versus 30 frames).} 
    \label{fig:bouncing_ball_results}
    \vspace{-2mm}
\end{figure}

{\bfseries \itshape Multiple objects with multiple dynamics.} We now turn from single-object sequences to sequences
involving multiple objects which operate according complex dynamics. We consider a bouncing-balls environment
in which multiple balls move with billiard-ball dynamics \citep{van2018relational}.  
The dataset consists of 50,000 training examples and 10,000 test examples showing $\sim$50 frames of either 4 solid 
same-color balls bouncing in a confined square geometry with different masses corresponding to their radii (\textit{4Balls}), 6-8 same-color balls  bouncing in a confined geometry (\textit{678Balls}), 3 same-color balls bouncing in a confined geometry with a central occluder (\textit{Curtain}), or balls of four different colors (\textit{Colored 678Balls}).  Although each ball has a distinct state (position, velocity, and possibly color), they share the same underlying dynamics of motion and collision. We expected that \modelname\ would learn to dissect these dynamics, e.g., 
by treating collisions, straight-line motion, interactions with the walls, and moving behind an occluder as distinct
schemata. We use the encoder and decoder architecture of \citep{van2018relational} for all models. \modelname\ has $n_f=4$
for 4Balls and Curtain, and $n_f=8$ for 678Balls and Coloured678Balls, and $n_s=4$ for all simulations. All models 
are trained for 100 epochs. As shown in Figure~\ref{fig:bouncing_ball_results}, \modelname\ achieves dramatic 
improvements in  successfully predicting 10- and 30-step iterated predictions relative to GRUs, and RIMs. 
(For further details, see Appendix Section \ref{sec:bouncing_ball}.) We omitted RMC \citep{santoro2018relational} 
from Figures~\ref{fig:bouncing_ball_results}a-d because RIMs performs strictly better than RMC in these 
environments \citep{goyal2019recurrent}, but have included RMC in the Appendix.


{\bfseries \itshape Increasing the number of schemata.} Our previous experiment used $n_s=4$ for all simulations. 
In order to study what happens if \modelname\ has access to a large number of schemata, we perform an experiment with
$n_s=10$. For the curtain task, we found that only three schemata were being used, and performance was about the same 
as when training with $n_s=3$ or $n_s=4$.  For 678Balls, we found that by the end of training only four schemas were
being used, and performance is about the same as when training with $n_s=4$. Thus, providing \modelname\ with
excess resources leads to some not being used which is a waste of computation and memory but does not lead to
suboptimal performance. For example, one might have been concerned about overfitting with too many resources.
However, with $n_s=3$ schemata, we never observe unused schemata, suggesting that the model does not have difficulty using the resources we provide to it. That is, there are no `dead' schemata that fail to be trained due to local optima.


\begin{table*}[tb!]
\centering
\scalebox{0.9}{
\begin{small}
\begin{tabular}{lcccccccccccc}
\toprule
&& \multicolumn{1}{c}{Block O1 (Disappear)} && \multicolumn{1}{c}{Block O2 (Shape Change)} && \multicolumn{1}{c}{Block O3 (Teleport)} \\
\toprule
RMC \citep{santoro2018relational} && 0.43 $\pm$ 0.05   &&  0.39 $\pm$ 0.03 &&  0.46 $\pm$ 0.02  \\
\midrule
Intphys \citep{riochet2019intphys}   && 0.52    &&  0.52  &&  0.51 \\
\midrule
\modelname\ (ours)  && \highlight{0.34 $\pm$ 0.05}   &&  \highlight{0.35 $\pm$ 0.05} &&  \highlight{0.42 $\pm$ 0.02} \\
\bottomrule
\end{tabular}
\end{small}}
\caption{{\it Results on the IntPhys benchmark.} Relative classification error of unrealistic physical phenomena \citep{riochet2019intphys} for three models, demonstrating  benefits of \modelname\  in scenes with significant occlusions ("Occluded"). This particular benchmark has three subsets, and for our experiments we evaluate the proposed model on the ``occlusion'' subset of the task.  The  three columns correspond to 3 different types of occlusions.  Lower is better. Average taken over 3 random seeds.}
\label{tab:intphys}
\end{table*}


{\bfseries \itshape Modeling physical laws in a multi-object environment.}
Modeling a physical system, such as objects in a world obeying laws of
gravity and momentum, requires factorization of state (object 
position, velocity) and dynamics. All objects must obey the same laws while
each object must maintain its distinct state.  We used the Intuitive Physics Benchmark \citep{riochet2019intphys} in which balls roll behind a brick wall such that they are briefly occluded. The training data is constructed such that the examples are all physically realistic.  The test set consists of both sequences that behave according to the laws used to synthesize the training data and sequences that follow unrealistic physical laws.
We test  three forms of unrealistic physics: balls disappearing behind the wall (O1 task), balls having their shape change for no reason (O2 task), and balls teleporting (O3 task). The Benchmark has three subsets of experiments, and we chose
the challenging subset with significant occlusions. We trained models to perform next state prediction on the training
set and we use model likelihood \citep{riochet2019intphys} to discriminate between realistic and unnatural test sequences. Further details are in Appendix~\ref{sec:intuitive_physics}.
As Table~\ref{tab:intphys} indicates,
\modelname\ significantly outperforms two competitors on all three tasks.


\section{Conclusions}

Understanding the visual world requires interpreting images in terms of distinct independent
physical entities. These entities have persistent intrinsic properties, such as a color or
velocity, and they have dynamics that transform the properties.  We explored a mechanism
that is able to factorize declarative knowledge (the properties) and procedural knowledge
(the dynamics).  Using attention, our \modelname\ model learns this factorization into
representations of entities---\OFs---and representations of how they transform over
time---schemata. By applying the same schemata to multiple \OFs, \modelname\ 
achieves systematicity of prediction, resulting in significantly improved generalization
performance over state-of-the-art methods. It also addresses a fundamental issue in
AI and cognitive science: the distinction between \emph{types} and \emph{tokens}. \modelname\ 
is also interpretable, in that
we can identify the binding between schemata and entity behavior. The factorization of
declarative and procedural knowledge has broad applicability to a wide variety of challenging
deep learning prediction tasks.

\section{Acknowledgements}
The authors acknowledge the important role played by their colleagues at Mila throughout the duration of this work. The authors  would like to thank Danilo Jimenez Rezende and Rosemary Nan Ke for useful discussions. The authors are grateful to Sjoerd van Steenkiste, Nicolas Chapados, Pierre-André Noël for useful feedback. The authors are grateful to NSERC, CIFAR, Google, Samsung, Nuance, IBM, Canada Research Chairs, Canada Graduate Scholarship Program, Nvidia for funding, and Compute Canada for computing resources.  We are very grateful to  Google for giving Google Cloud credits used in this project.

\bibliographystyle{plainnat}
\bibliography{main}
\clearpage

\appendix
\onecolumn


\addcontentsline{toc}{section}{Appendix} 
\part{Appendix} 
\parttoc 

\section{Implementation Details and Hyperparameters}
\label{sec:hyperparameters}
The  model setup consists of three main components: an encoder, the process of interaction between object files and schemata followed by a decoder. The images are first processed by an encoder, which is parameterized as a CNN.

\textbf{Resources Used.} It takes about 2 days to train the proposed model on bouncing ball task for 100 epochs on V100 (32G). We did not do any hyper-parameter search specific to a particular dataset (i.e 4Balls or 678Balls or Curtain Task). We ran the proposed model for different number of schemata (i.e 2/4/6). Similarly, it takes about 3 days to run for 20M steps for the Reinforcement learning task. 



\section{Adding Task}

\begin{figure}
    \centering
    \subfigure[]{\includegraphics[width=0.52\textwidth, height= 0.19\textwidth]{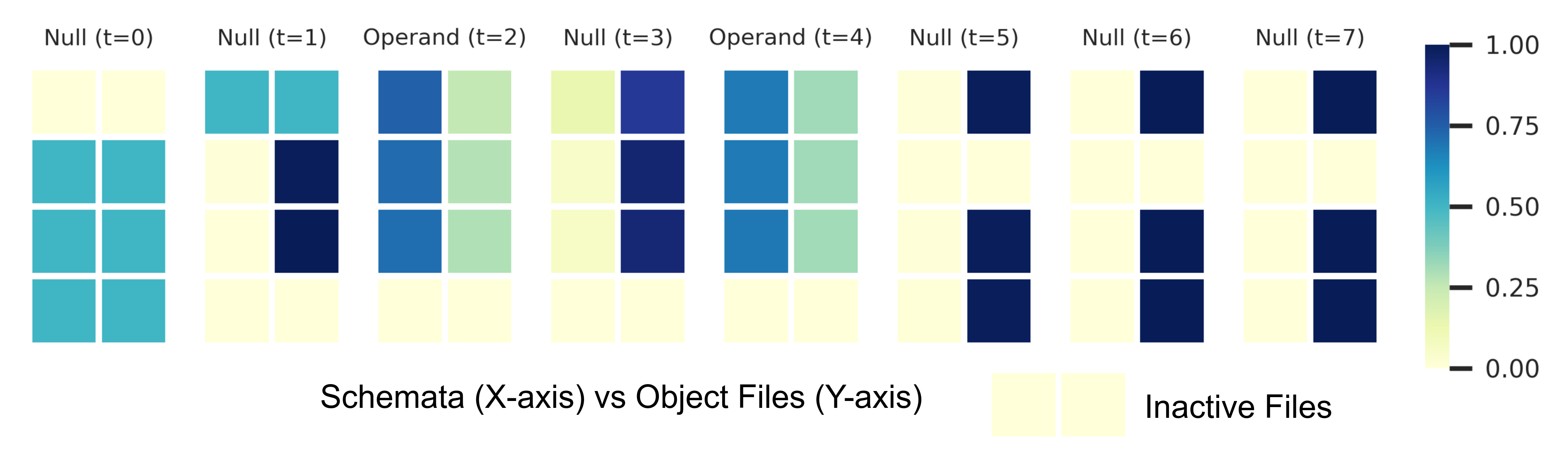}} 
    \subfigure[]{\includegraphics[width=0.45\textwidth]{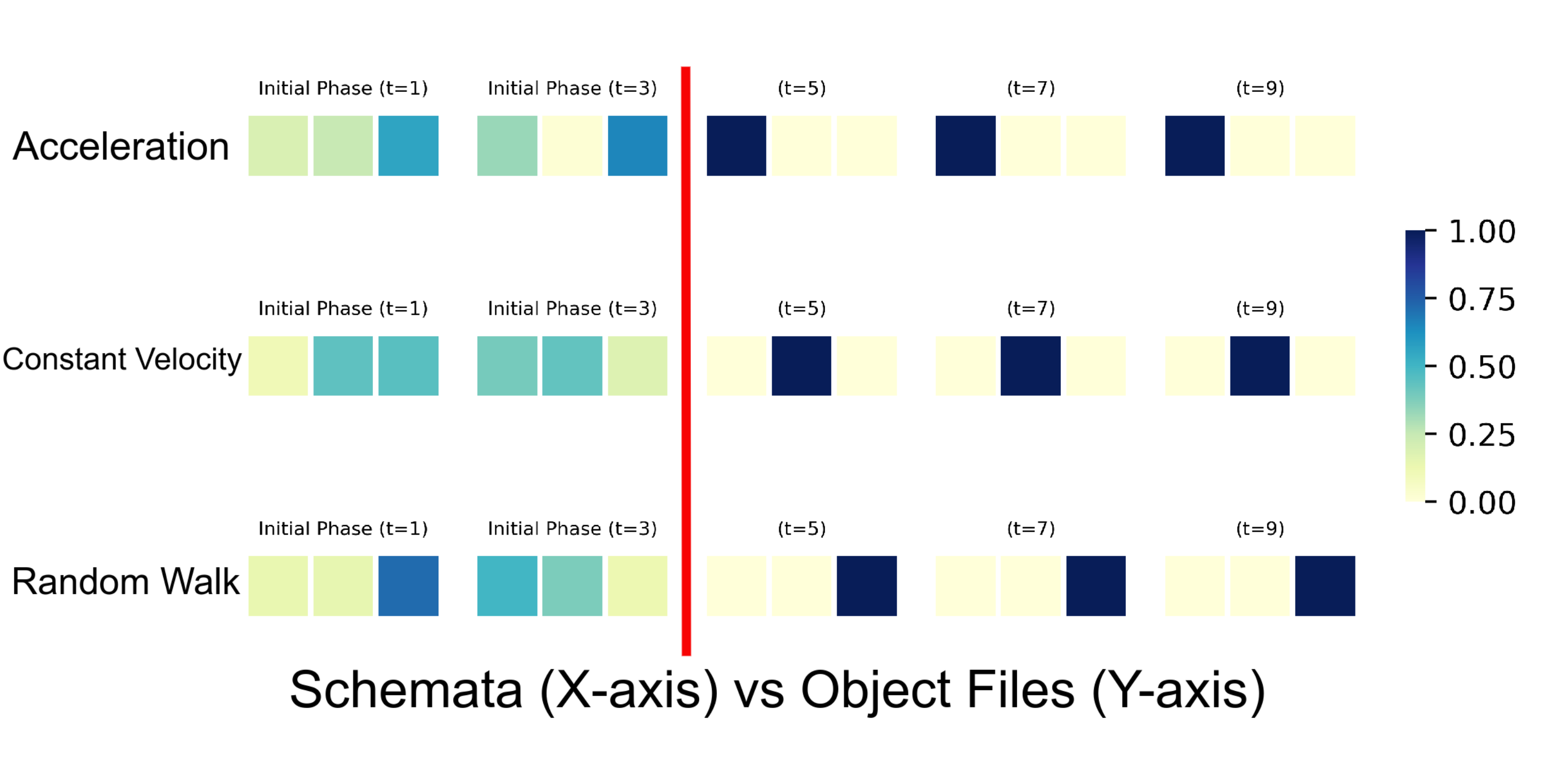}} 
\    \caption{(a) \OF\ ($n_f = 4$) vs Schemata ($n_s = 2$) activation for an example of length 8 of the adding task. "Null" refers to the elements other than the operands on which the addition is to be performed. The figure shows the affinity of each \OF\ to use a particular schema. Each row corresponds to a particular \OF, and column represents a particular schema (dark color shows high affinity of an \OF\  toward a particular schema). As shown in the figure, the active \OFs\ trigger Schema 1 when an operand is encountered, and Schema 2 when a "Null" element is encountered. (b) Here, we have a single \OF, and that can follow three different dynamics. We found that our method is able to learn these 3 different modes once it's passed an initial phase of uncertainty.}
     \label{fig:adding_appendix}
\end{figure}

We analyzed the proposed method on the adding task. This is a standard task for investigating recurrent models \citep{hochreiter1997long}. The input consists of two co-occuring sequences: 1) N numbers $(a_{0} \cdots a_{N−1})$ sampled independently from $U[0, 1]$, 2) an index  $i_{0}$ in the first half of the sequence, and an index $i_{1}$ in the second half of the sequence together encoder as a one hot sequences.  The target output is $a_{i_{0}}$ + $a_{i_{1}}$.  As shown in figure~\ref{fig:adding_appendix}~(a), we can clearly observe the factorisation of procedural knowledge into two schemata effectively, one schema is triggered when an operand is encountered and the other when non-operand is encountered. 

\paragraph{Generalization Result:}
For demonstrating the generalization capability of \modelname, we consider a scenario where the models are trained to add a mixture of two and four numbers from sequences of length 50. They are evaluated on adding variable number (2-10) of numbers on 200 length sequences. As shown in Table 1, we note better  generalization when using \modelname. The dataset consists of 50,000 training sequences and 20,000 testing sequences for each different number of numbers to add. 

\label{sec:adding_task}
All the models are trained for 100 epochs with a learning rate of 0.001 using the Adam optimizer. We use 300 as the hidden dimension for both the LSTM baseline and the LSTM's in RIMS, \modelname. Table \ref{table::appendix::adding::hyperparameters} lists the different hyperparameters used for training \modelname.
\begin{table}[tbh]
    \caption{Hyperparameters for the adding generalization task}
  \begin{center}
    \begin{tabular}{lr}
      \toprule
      Parameter & Value  \\
      \midrule
      Number of object files ($n_f$)
& 5 \\

Number of schemata ($n_s$)
& 2 \\
   Optimizer                                   & Adam\citep{Kingma2014}\\
      learning rate                                   & $1\cdot 10^{-2}$      \\
      batch size                                      & 64      \\
     Inp keys &  64 \\
     Inp Values & 60 \\
     Inp Heads & 4 \\ 
     Inp Dropout & 0.1 \\
     Comm keys &  32 \\
     Comm Values & 32 \\
     Comm heads & 4 \\ 
     Comm Dropout & 0.1 \\
      \bottomrule
    \end{tabular}
  \end{center}
  \label{table::appendix::adding::hyperparameters}
\end{table}

\begin{table}
    \centering
    \scalebox{0.90}{
    \begin{tabular}{c|c c c}
    \toprule
    \textit{Number of Values} & \textit{LSTM}  & \textit{RIMS} & \textit{\modelname} \\
    \midrule
    2& 0.8731 & 0.0007 & \highlight{0.0005}\\
    3& 1.3017 & 0.0009 & \highlight{0.0007}\\
    4& 1.6789 & 0.0014 & \highlight{0.0013}\\
    5& 2.0334 & 0.0045 & \highlight{0.0030}\\
    8& 4.8872 &0.0555 & \highlight{ 0.0191}\\
    9& 7.3730 & 0.1958 & \highlight{0.0379}\\
    10& 11.3595 & 0.8904 & \highlight{0.0539}\\
    \bottomrule
      
    \end{tabular}
   
    }
   
     \caption{\textbf{Adding Task: }Mean test set error on 200 length sequences with number of numbers to add varying among $\{2,3,4,5,8,9,10\}$. The models are trained to add a mixture of two and four numbers from sequences of length 50.}
    
     \hspace{1.0pt}
 \label{tab:adding_generalization}
\vspace{-4mm}
\end{table}

\section{Bouncing Ball}
The dataset consists of 50,000 training examples and 10,000 test examples showing $\sim$50 frames of either 4 solid balls bouncing in a confined square geometry (\textit{4Balls}), 6-8 balls bouncing in a confined geometry (\textit{678Balls}), 3 balls bouncing in a confined geometry with an occluded region (\textit{Curtain}), or balls of different colors (\textit{Colored 678Balls}).  The balls have distinct states (and hence distinct object files) but share underlying procedures (schema), which we aim to capture using \modelname.  


We trained baselines as well as proposed model for about 100 epochs.  We also provide the rollouts predicted by the models in the Figures \ref{fig:rollouts_4balls}, \ref{fig:rollouts_678balls}, \ref{fig:rollouts_curtain}, \ref{fig:rollouts_color4balls}, \ref{fig:rollouts_color678balls}. 

As part of future work, We also provide a scenario where instead of activating all the \OF\, only a subset of the \OF\ that are most relevant to the input based on attention scores are activated. We denote the original model as \modelname-ab, and the ablation of only activating a subset of the \OF\ as  \modelname.

\begin{figure}[bt]
    \centering
    
    \subfigure[\textbf{Coloured678Balls}]{\includegraphics[ width=0.7\textwidth,trim={2.5cm 1.3cm 3cm 1cm},clip]{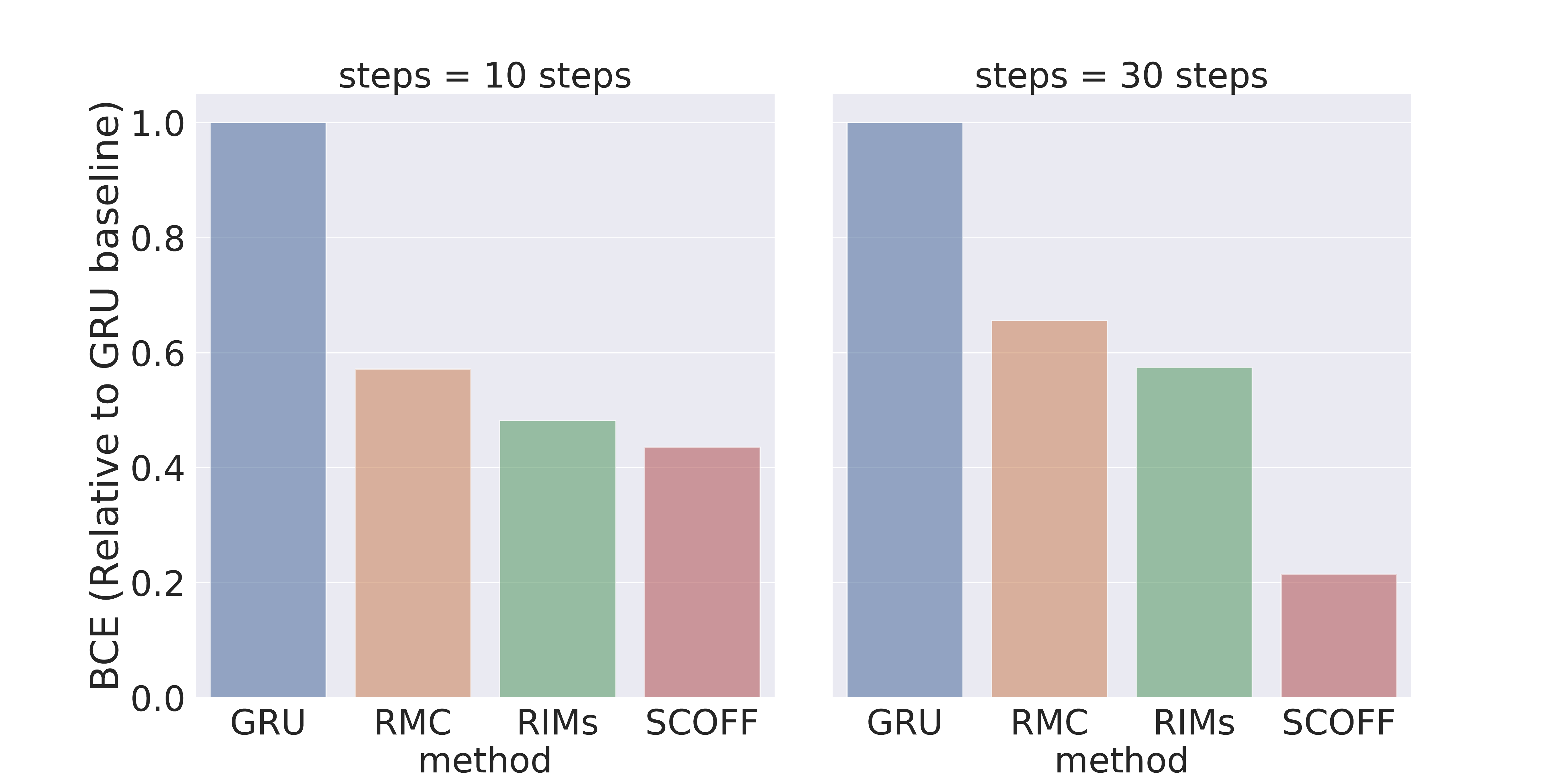}}\par

    \caption{\textit{Bouncing ball motion.} Error relative to GRU baseline on 10- and 30-frame video prediction
    of multiple-object video sequences  Predictions are based on 15 frames of ground truth.} 
    \label{fig:bouncing_ball_results_rmc}
    \vspace{-2mm}
\end{figure}

Comparison to the RMC baseline is shown in fig. \ref{fig:bouncing_ball_results_rmc}.

\label{sec:bouncing_ball}
\begin{table}[tbh]
    \caption{Hyperparameters for the bouncing balls task}
  \begin{center}
    \begin{tabular}{lr}
      \toprule
      Parameter & Value  \\
      \midrule
      Number of object files ($n_f$)
& 4\\

Number of schemata ($n_s$) & 2/4/6\\
Size of Hidden state of  object file & 100 \\
   Optimizer                                   & Adam\citep{Kingma2014}\\
      learning rate                                   & $1\cdot 10^{-4}$      \\
      batch size                                      & 64      \\
     Inp keys &  64 \\
     Inp Values & 100 \\
     Inp Heads & 1 \\ 
     Inp Dropout & 0.1 \\
     Comm keys &  32 \\
     Comm Values & 32 \\
     Comm heads & 4 \\ 
     Comm Dropout & 0.1 \\
      \bottomrule
    \end{tabular}
  \end{center}
  \label{table::appendix::bouncing balls::hyperparameters}
\end{table}

\begin{figure}[h]
    \centering
    \includegraphics[width=0.9\textwidth , height= 0.9\textheight,keepaspectratio ]{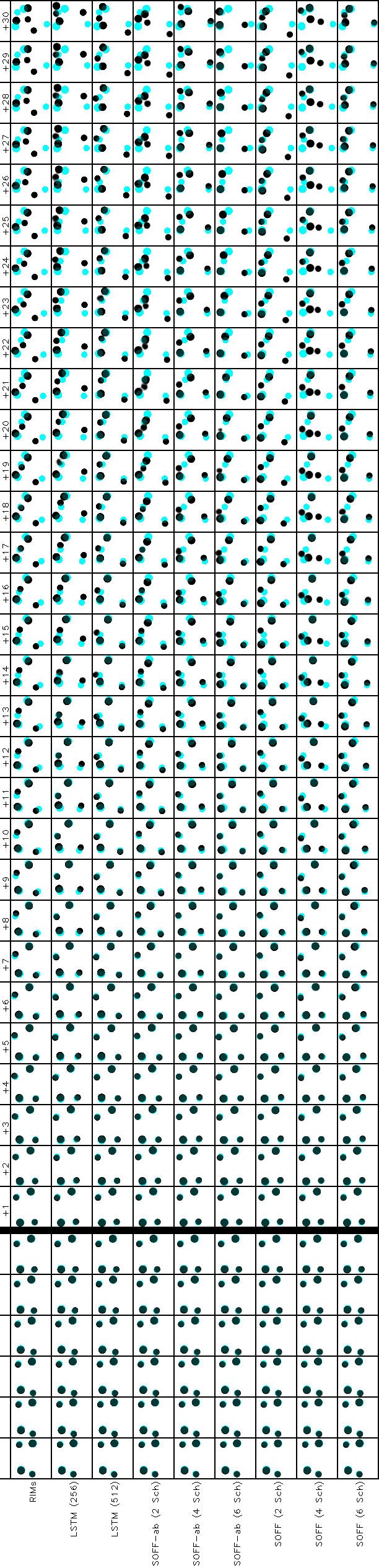}
  
    \caption{\textbf{Rollout for 4Balls.} In all cases, the first 10 frames of ground truth are fed in (last 6 shown) and then the
system is rolled out for the next 30 time steps. In the predictions, the transparent blue shows the ground truth, overlaid to help guide the eye. }
    \label{fig:rollouts_4balls}
\end{figure}

\begin{figure}[h]
    \centering
    \includegraphics[width=0.9\textwidth , height=0.9\textheight,keepaspectratio]{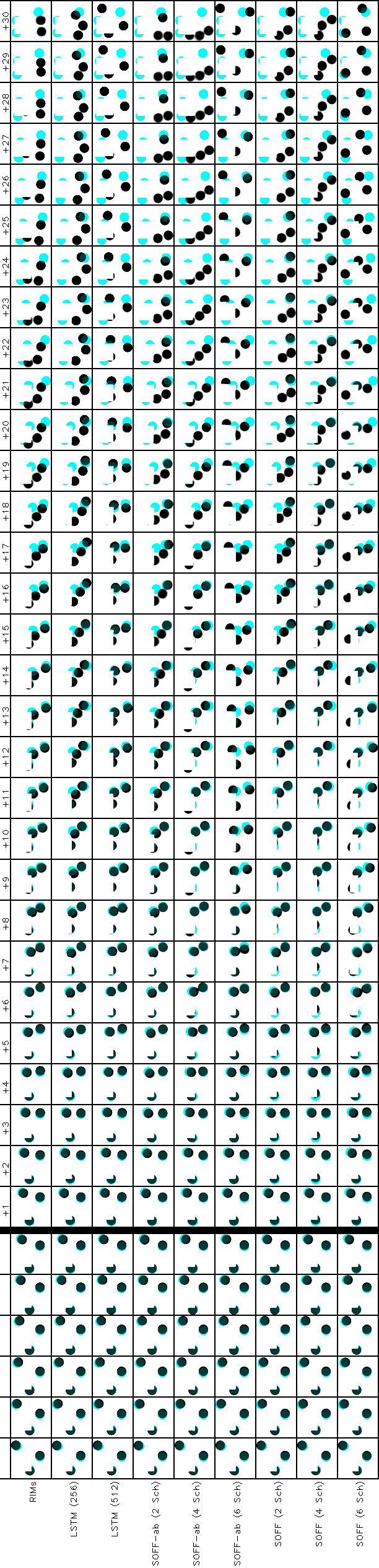}
    \caption{\textbf{Rollout for Curtain.} In all cases, the first 10 frames of ground truth are fed in (last 6 shown) and then the
system is rolled out for the next 30 time steps. In the predictions, the transparent blue shows the ground truth, overlaid to help guide the eye.}
    \label{fig:rollouts_curtain}
\end{figure}

\begin{figure}[h]
    \centering
    \includegraphics[width=0.9\textwidth , height= 0.9\textheight,keepaspectratio]{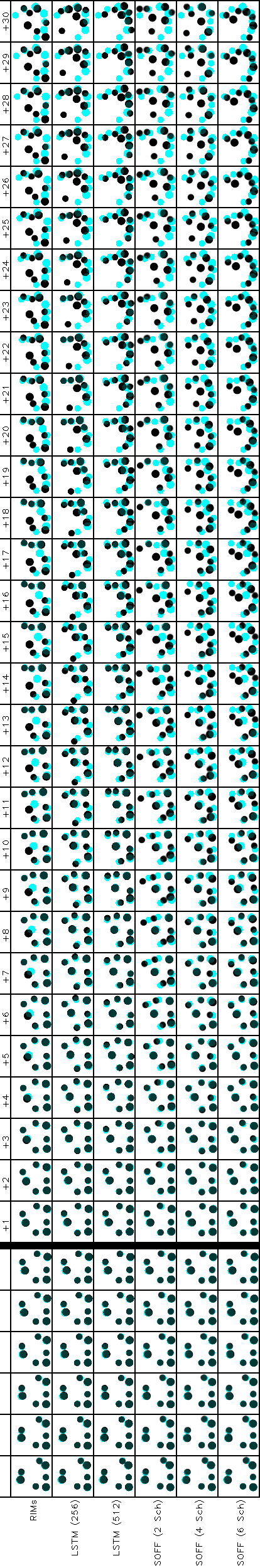}
    \caption{\textbf{Rollout for 678Balls.} In all cases, the first 10 frames of ground truth are fed in (last 6 shown) and then the
system is rolled out for the next 30 time steps. In the predictions, the transparent blue shows the ground truth, overlaid to help guide the eye.}
    \label{fig:rollouts_678balls}
\end{figure}

\begin{figure}[h]
    \centering
    \includegraphics[width=0.9\textwidth , height= 0.9\textheight,keepaspectratio]{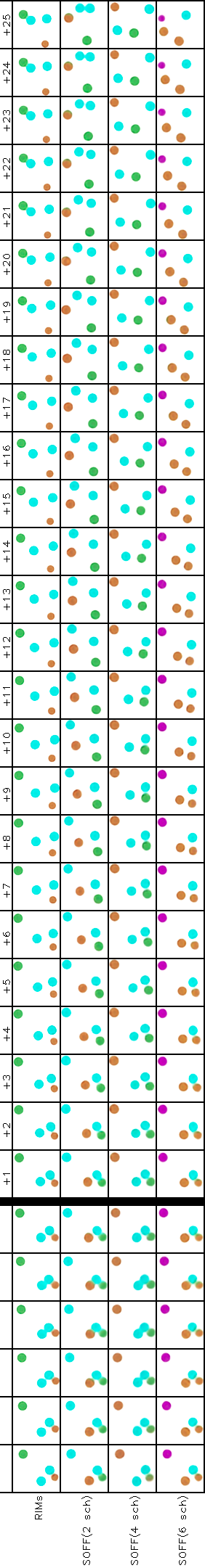}
    \caption{\textbf{Rollout for Colored 4Balls.} In all cases, the first 10 frames of ground truth are fed in (last 6 shown) and then the
system is rolled out for the next 25 time steps.}
    \label{fig:rollouts_color4balls}
\end{figure}

\begin{figure}[h]
    \centering
    \includegraphics[width=0.9\textwidth , height= 0.9\textheight,keepaspectratio]{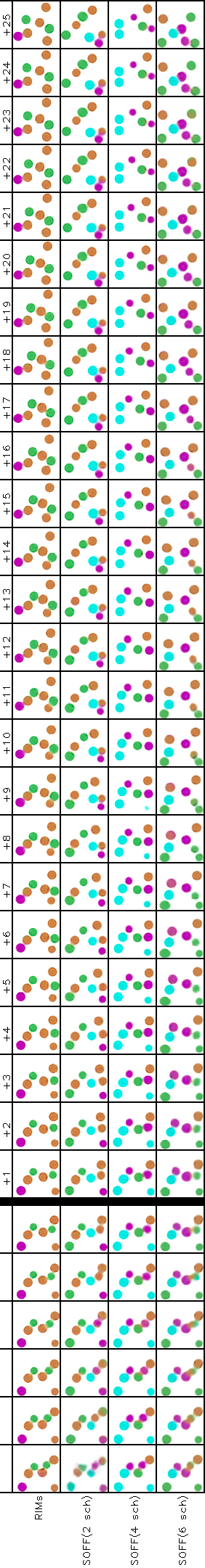}
    \caption{\textbf{Rollout for Colored 678Balls.} In all cases, the first 10 frames of ground truth are fed in (last 6 shown) and then the
system is rolled out for the next 25 time steps.}
    \label{fig:rollouts_color678balls}
\end{figure}

\section{BabyAI: Reinforcement Learning}
\label{sec:reinforcement_learning}

We use the GotoObjMaze environment (i.e MiniGrid-GotoObjMaze-v0) from \citep{chevalier2018babyai}.  Here, the agent has to navigate to an object, the object may be in another room. We use  exactly the same RL setup as in \citep{chevalier2018babyai} except we extend the setup in BabyAI to only apply RGB images of the world rather than symbolic state representations, and hence making the task much more difficult.  Hyper-parameters for this task are listed in Tab. \ref{table::appendix::babyai::hyperparameters}. 

\begin{figure}
    \centering
    \subfigure[Object Files ($n_f=1$) (vs) Schemata ($n_s=4$) Activation]{\includegraphics[width=0.7\textwidth,height= 0.3\textheight, keepaspectratio]{iclr2021_results/rl_trajectory.pdf}}\par
        \subfigure[Agent view in the environment]{\includegraphics[width=0.9\textwidth,height= 0.5\textheight,keepaspectratio]{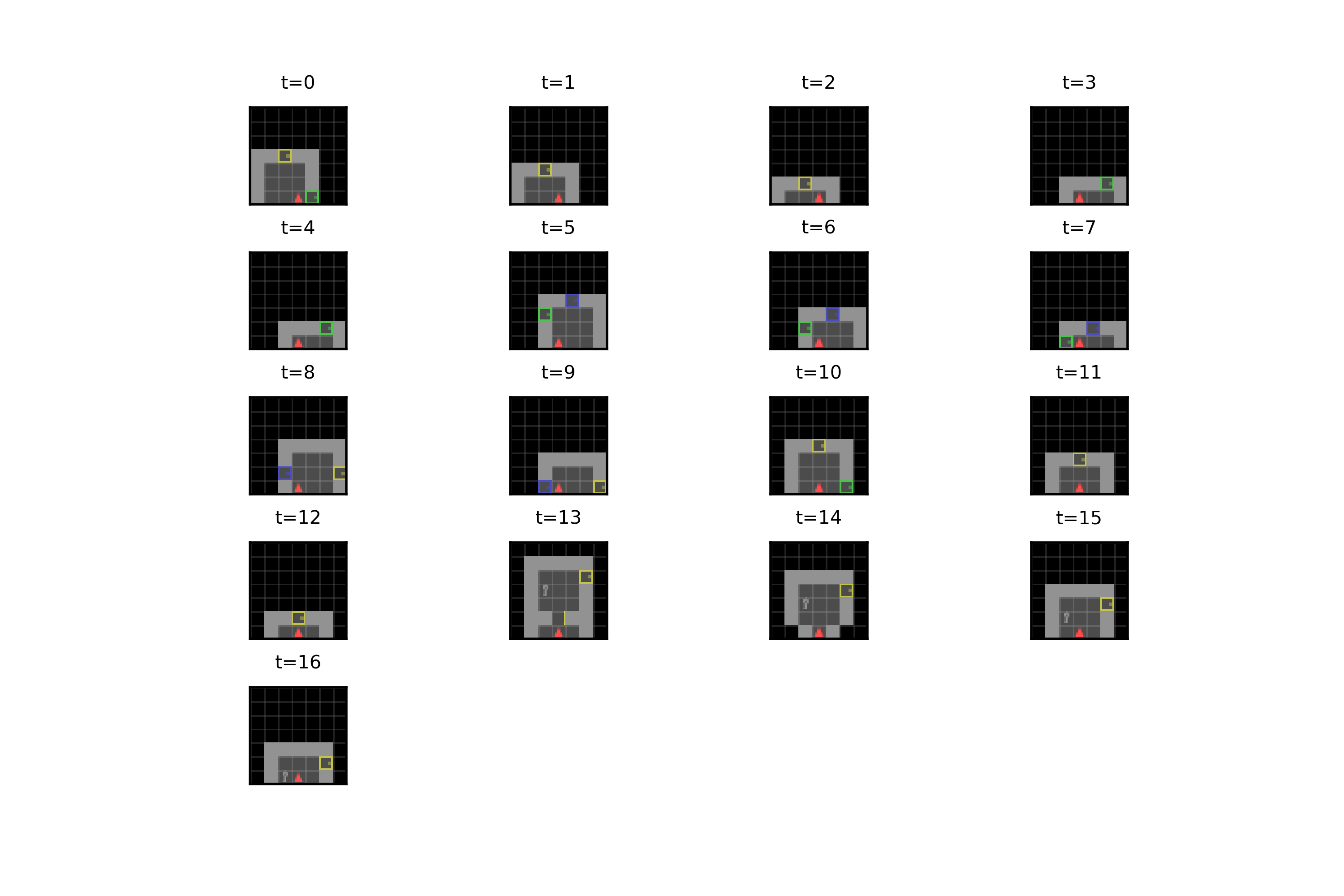}}
    \caption{\textbf{BabyAI-GotoObjMaze Trajectory} In this environment, the agent is expected to navigate a 3x3 maze of 6x6 rooms, randomly inter-connected
by doors to find an object like "key". Here we use only one object file, but different number of schemata (4 in this example). If we look at the object files (vs) schemata affinity, schema 1 is activated while close to or opening doors while schema 4 is  triggered when the "key" is in the agent's view.}
    \label{fig:babyai}
\end{figure}

In this environment, the agent is expected to navigate a 3x3 maze of 6x6 rooms, randomly inter-connected by doors to find an object like "key". Here we use only one object file, but different number of schemas (4 in this example). If we look at the object files (vs) schemata, schemata 4 is being triggered when the "key" is in agent's view as shown in fig. \ref{fig:babyai}.

\begin{table}[tbh]
    \caption{Hyperparameters for BabyAI}
  \begin{center}
    \begin{tabular}{lr}
      \toprule
      Parameter & Value  \\
      \midrule
Number of object files ($n_f$) & 1 \\
Number of schemata ($n_s$) &  2/4/6 \\
Size of Hidden state of object file & 510 \\
  Optimizer                 & Adam\citep{Kingma2014}\\
     Learning rate          & $3\cdot 10^{-4}$      \\
     Inp keys &  64 \\
     Inp Values & 256 \\
     Inp Heads & 4 \\ 
     Inp Dropout & 0.1 \\
     Comm keys &  16 \\
     Comm Values & 32 \\
     Comm heads & 4 \\ 
     Comm Dropout & 0.1 \\
      \bottomrule
    \end{tabular}
  \end{center}
  \label{table::appendix::babyai::hyperparameters}
\end{table}

\section{Intuitive Physics}
\label{sec:intuitive_physics}

\begin{table}[tbh]
    \caption{Hyperparameters for IntPhys benchmark}
  \begin{center}
    \begin{tabular}{lr}
      \toprule
      Parameter & Value  \\
      \midrule
      Number of object files ($n_f$) &  6 \\
      Number of schemata ($n_s$) & 4/6 \\
      Optimizer         & Adam\citep{Kingma2014}\\
      learning rate                                   & $3\cdot 10^{-4}$      \\
      batch size                                      & 64      \\
     Inp keys &  64 \\
     Inp Values & 85 \\
     Inp Heads & 4 \\ 
     Inp Dropout & 0.1 \\
     Comm keys &  32 \\
     Comm Values & 32 \\
     Comm heads & 4 \\ 
     Comm Dropout & 0.1 \\
      \bottomrule
    \end{tabular}
  \end{center}
  \label{table::appendix::physics::hyperparameters}
\end{table}

We use the similar training setup as \citep{riochet2019intphys}. Hyper-parameters related to the proposed method are listed in Tab. \ref{table::appendix::physics::hyperparameters}.

\section{Switching Dynamics}
\begin{figure}[h]
    \centering
    \includegraphics[width=0.9\textwidth , height= 0.9\textheight,keepaspectratio]{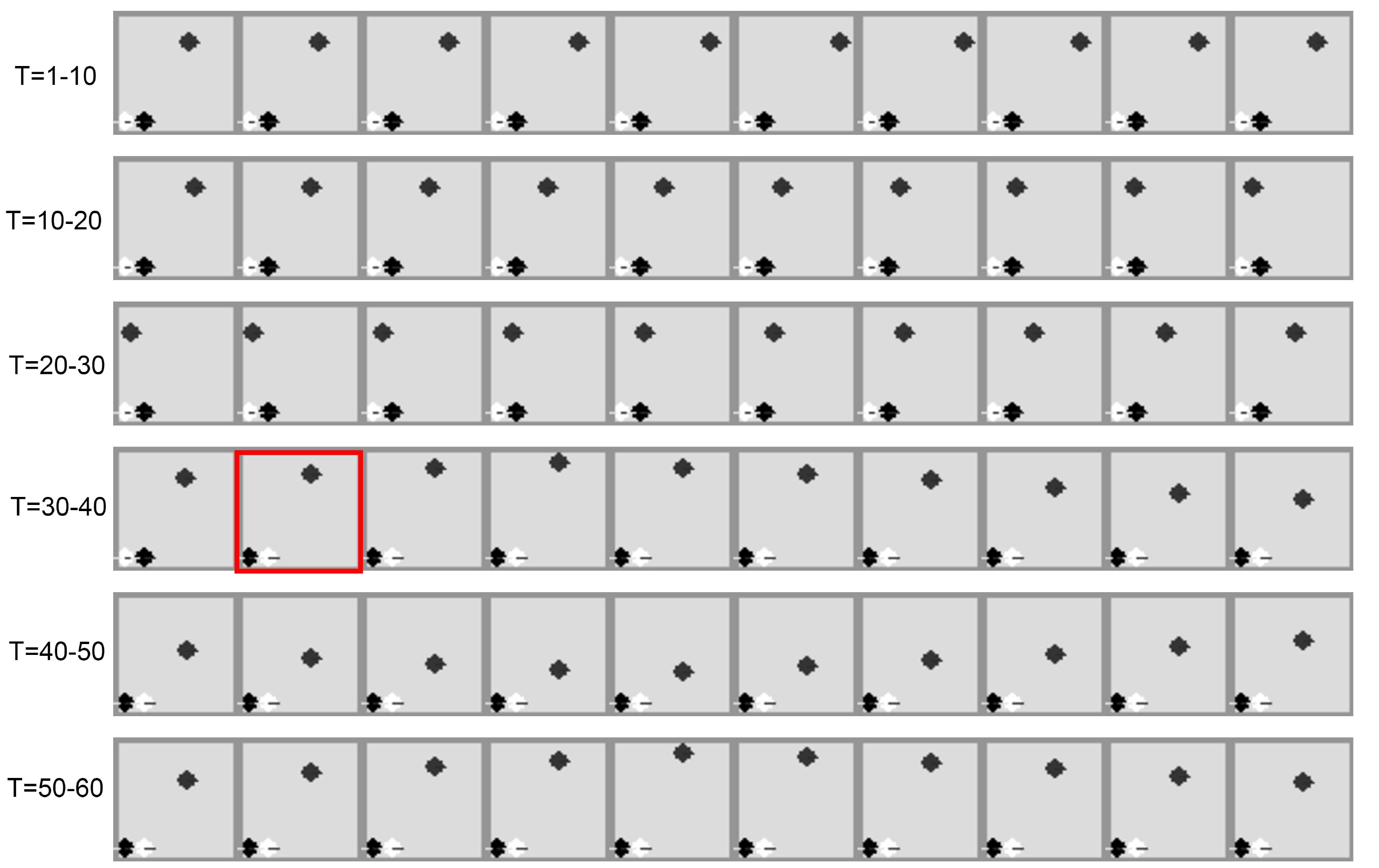}
    \caption{\textbf{Switching dynamics task.} An example ground truth trajectory, where the ball oscillates horizontally  and switches to vertical oscillations  after few steps indicated by the red box. The bottom left has two bulb indicators corresponding to the two dynamics.}
    \label{fig:dynamics_gt}
\end{figure}

We consider a scenario where a ball follows one of the two dynamics, horizontal and vertical oscillations at a given time. We limit the  number of switches between the dynamics followed to be one. An example ground truth trajectory is given in figure~\ref{fig:dynamics_gt}. Here we run another experiment, where we have more number of \OFs\ as compared to number of entities or objects, and then we investigate if we are still able to have factorization of different dynamics in different schemata. 

\subsection{Experiment Setup}
The dataset consists of 10,000 trajectories of 51 length with the switching between dynamics happening at the middle of the trajectory. 

We use the same architecture for encoder as well as decoder as in \citep{van2018relational}. At each time step, we give the last five frames stacked across the channels as input to the encoder. 

\begin{table}[tbh]
    \caption{Hyperparameters for the switching  dynamics task}
  \begin{center}
    \begin{tabular}{lr}
      \toprule
      Parameter & Value  \\
      \midrule
      Number of object files ($n_f$) &  6 \\
      Number of schemata ($n_s$) & 2 \\
      Optimizer         & Adam\citep{Kingma2014}\\
      learning rate                                   & $1\cdot 10^{-4}$      \\
      batch size                                      & 50      \\
     Inp keys &  64 \\
     Inp Values & 64 \\
     Inp Heads & 4 \\ 
     Inp Dropout & 0.1 \\
     Comm keys &  32 \\
     Comm Values & 32 \\
     Comm heads & 4 \\ 
     Comm Dropout & 0.1 \\
      \bottomrule
    \end{tabular}
  \end{center}
  \label{table::appendix::dynamics::hyperparameters}
\end{table}

\subsection{The factorisation of dynamics into schemata}
\begin{figure}[h]
    \centering
    \subfigure[Object Files ($n_f=6$) (vs) Schemata ($n_s=2$) for trajectory one.]{\includegraphics[width=0.98\textwidth , height= 0.43\textheight, keepaspectratio]{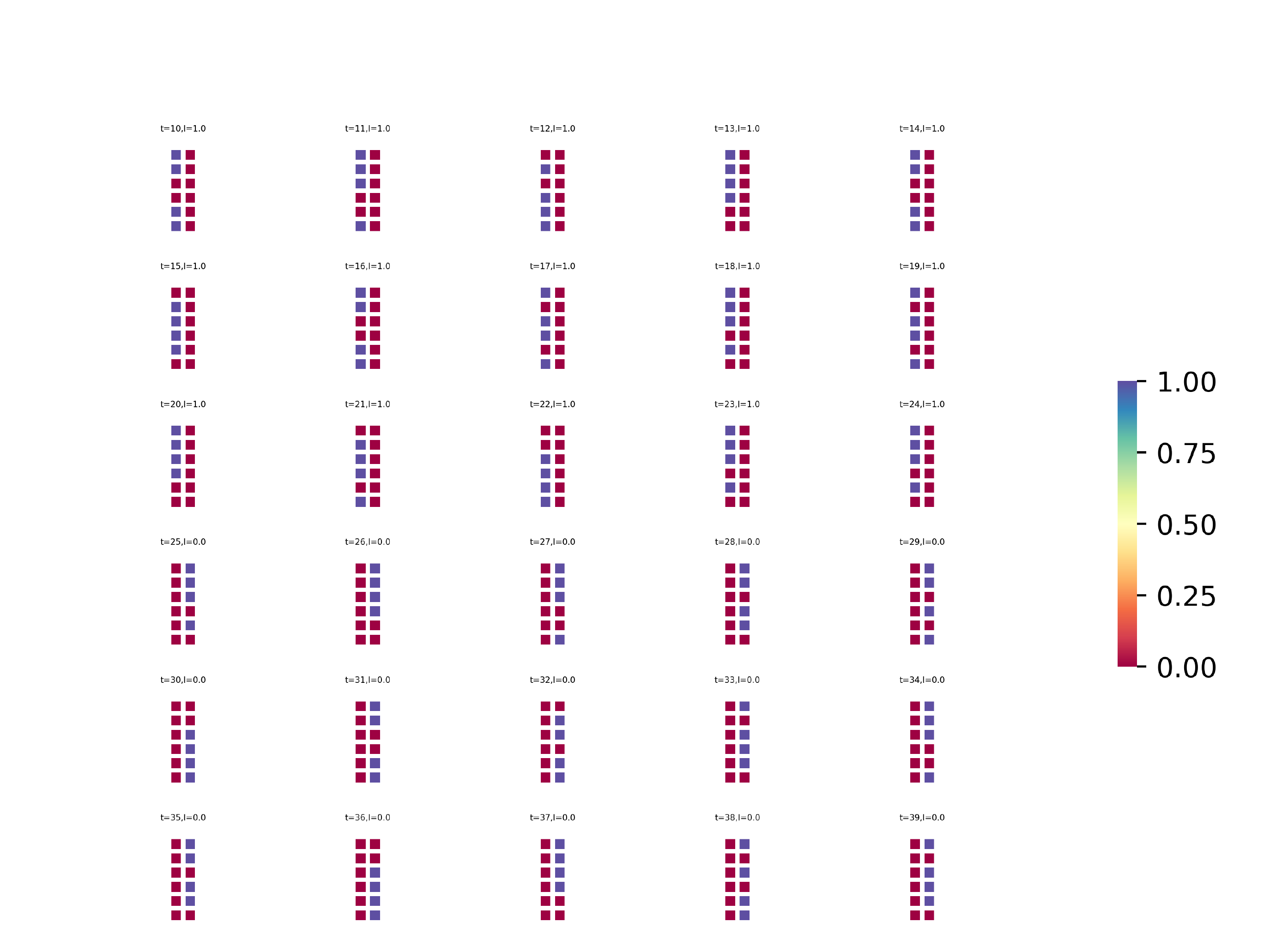}} \\
        \subfigure[Object Files ($n_f=6$) (vs) Schemata ($n_s=2$) for trajectory two.]{\includegraphics[width=0.98\textwidth , height= 0.43\textheight, keepaspectratio]{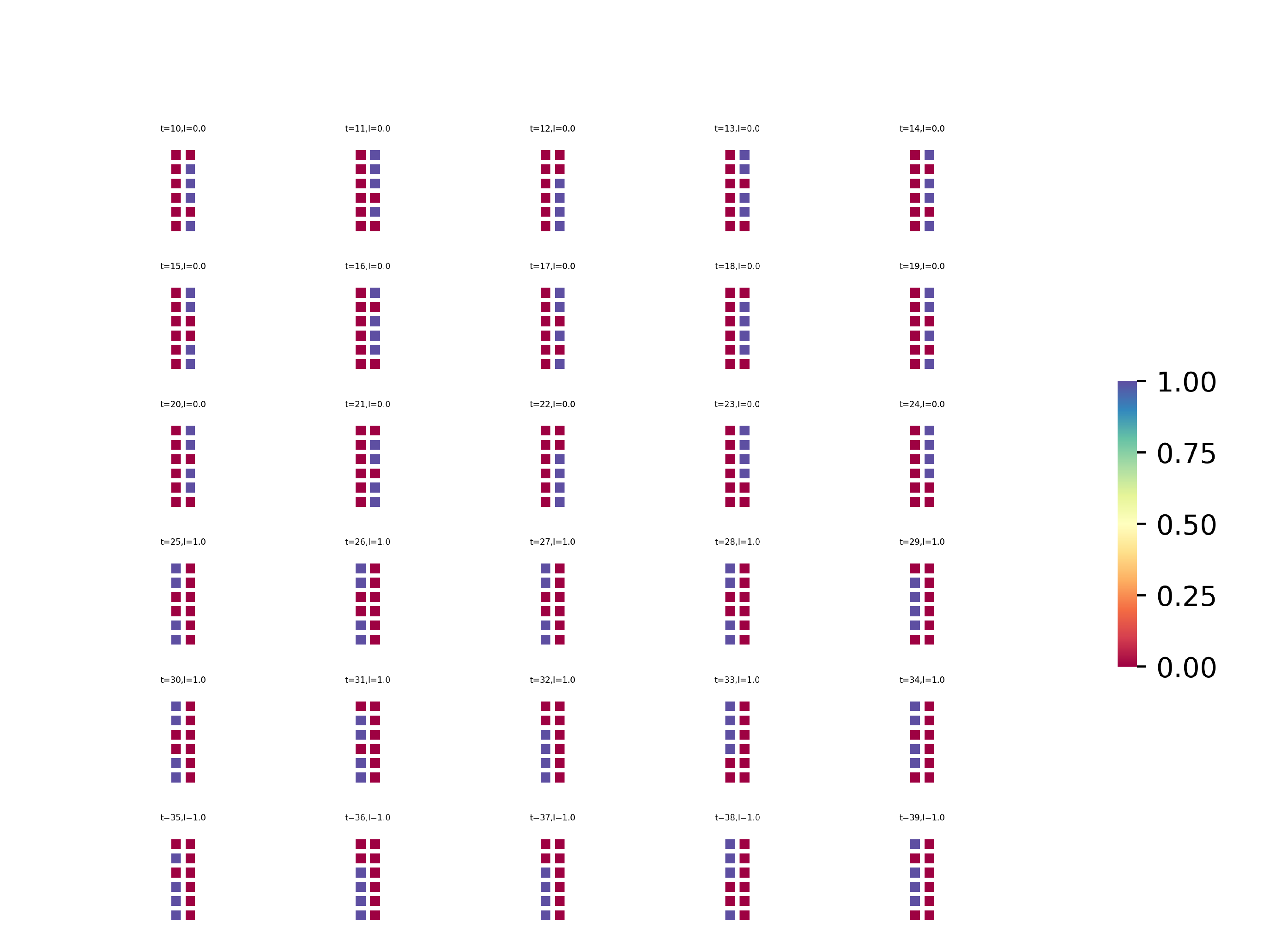}} 
    \caption{\textbf{Switching dynamics task.} $l=0$ denotes horizontal oscillations and $l=1$ denotes vertical oscillations. We can clearly observe that the dynamics are being factorised into separate schemata. Schemata one is being used for vertical oscillations and schemata two for horizontal oscillations. }
    \label{fig:dynamics}
\end{figure}

As shown in figure~\ref{fig:dynamics}, \modelname~ is effective in factorising the two different dynamics into two different schemata, even if the order of dynamics followed is different.

\end{document}